\documentclass[10pt,twocolumn,letterpaper]{article}

\usepackage{cvpr}
\usepackage{times}
\usepackage{epsfig}
\usepackage{graphicx}
\usepackage{amsmath}
\usepackage{amssymb}

\usepackage[utf8]{inputenc} 
\usepackage[T1]{fontenc}    
\usepackage{url}            
\usepackage{booktabs}       
\usepackage{amsfonts}       
\usepackage{nicefrac}       
\usepackage{microtype}      
\usepackage{latexsym}
\usepackage{comment}        
\usepackage{subfigure}
\usepackage{multirow}

\usepackage{bm}
\usepackage{amsthm}
\usepackage{makecell}
\usepackage{mathtools}
\usepackage{color}
\usepackage{xcolor}
\usepackage{MnSymbol}
\usepackage{makecell}
\usepackage{arydshln}
\usepackage[shortlabels]{enumitem}
\usepackage{booktabs}
\usepackage[font=small]{caption}
\usepackage[scientific-notation=true]{siunitx}
\usepackage{wrapfig}
\usepackage{lipsum}
\usepackage{sidecap}
\usepackage{pifont}

\newcommand{\namel}{Cross-modal Meta-Alignment}
\newcommand{\names}{\textsc{CroMA}}
\newcommand{\name}{\textsc{CroMA}}

\definecolor{gg}{RGB}{15,125,15}
\definecolor{rr}{RGB}{190,45,45}

\newcommand{\PreserveBackslash}[1]{\let\temp=\\#1\let\\=\temp}
\newcolumntype{C}[1]{>{\PreserveBackslash\centering}p{#1}}
\newcolumntype{R}[1]{>{\PreserveBackslash\raggedleft}p{#1}}
\newcolumntype{L}[1]{>{\PreserveBackslash\raggedright}p{#1}}

\newtheorem{proposition}{Proposition}
\newtheorem{assumption}{Assumption}
\newtheorem{corollary}{Corollary}
\newtheorem{definition}{Definition}

\usepackage{algorithm}
\usepackage[noend]{algpseudocode}
\algdef{SE}[SUBALG]{Indent}{EndIndent}{}{\algorithmicend\ }%
\algtext*{Indent}
\algtext*{EndIndent}

\makeatletter
\newcommand*\wt[1]{\mathpalette\wthelper{#1}}
\newcommand*\wthelper[2]{%
        \hbox{\dimen@\accentfontxheight#1%
                \accentfontxheight#11.3\dimen@
                $\m@th#1\widetilde{#2}$%
                \accentfontxheight#1\dimen@
        }%
}

\newcommand*\accentfontxheight[1]{%
        \fontdimen5\ifx#1\displaystyle
                \textfont
        \else\ifx#1\textstyle
                \textfont
        \else\ifx#1\scriptstyle
                \scriptfont
        \else
                \scriptscriptfont
        \fi\fi\fi3
}
\makeatother

\DeclareMathOperator*{\argmin}{arg\,min}
\DeclareMathOperator*{\argmax}{arg\,max}

\usepackage[pagebackref=true,breaklinks=true,letterpaper=true,colorlinks,bookmarks=false]{hyperref}

\cvprfinalcopy 

\def\cvprPaperID{6796} 

\ifcvprfinal\fi
\begin{document}

\title{Cross-Modal Generalization:\\Learning in Low Resource Modalities via Meta-Alignment}

\author{%
  Paul Pu Liang$^\spadesuit$, Peter Wu$^\spadesuit$, Liu Ziyin$^\heartsuit$, Louis-Philippe Morency$^\spadesuit$, Ruslan Salakhutdinov$^\spadesuit$\\
  $^\spadesuit$Carnegie Mellon University, $^\heartsuit$University of Tokyo\\
  \texttt{\{pliang,peterw1\}@cs.cmu.edu}\\
}

\maketitle

\begin{abstract}
The natural world is abundant with concepts expressed via visual, acoustic, tactile, and linguistic modalities. Much of the existing progress in multimodal learning, however, focuses primarily on problems where the same set of modalities are present at train and test time, which makes learning in low-resource modalities particularly difficult. In this work, we propose algorithms for cross-modal generalization: a learning paradigm to train a model that can (1) quickly perform new tasks in a target modality (i.e. meta-learning) and (2) doing so while being trained on a different source modality. We study a key research question: how can we ensure generalization across modalities despite using separate encoders for different source and target modalities? Our solution is based on meta-alignment, a novel method to align representation spaces using strongly and weakly paired cross-modal data while ensuring quick generalization to new tasks across different modalities. We study this problem on 3 classification tasks: text to image, image to audio, and text to speech. Our results demonstrate strong performance even when the new target modality has only a few (1-10) labeled samples and in the presence of noisy labels, a scenario particularly prevalent in low-resource modalities.
\end{abstract}

\vspace{-6mm}
\section{Introduction}
\vspace{-1mm}

One of the hallmarks of human intelligence is the ability to generalize seamlessly across heterogeneous sensory inputs and different cognitive tasks~\cite{DBLP:journals/corr/abs-1911-01547}. We see objects, hear sounds, feel textures, smell odors, and taste flavors to learn underlying concepts present in our world~\cite{baltruvsaitis2018multimodal}. Much of AI's existing progress in multimodal learning, however, focuses primarily on a fixed set of predefined modalities and tasks~\cite{DBLP:journals/corr/KrishnaZGJHKCKL16,liang2019learning} that are consistent between training and testing. As a result, it is unclear how to transfer knowledge from models trained for one modality (e.g. visual source modality) to another (e.g. audio target modality) at test time.
This scenario is particularly important for low-resource target modalities where unlabeled data is scarce and labeled data is even harder to obtain (e.g. audio from low-resource languages~\cite{li2019santlr}, real-world environments~\cite{DBLP:journals/corr/abs-1808-00177}, and medical images~\cite{DBLP:journals/corr/abs-1903-11101}). In the unimodal case, this is regarded as meta-learning~\cite{finn2017model} or few-shot learning~\cite{chen2019closer}.
In contrast, we formally define the \textit{cross-modal generalization} setting as a learning paradigm to train a model that can (1) quickly perform new tasks in a target modality (i.e. meta-learning) and (2) doing so while being trained on a different source modality.
In this paper, we study the data and algorithmic challenges for cross-modal generalization to succeed.

\begin{figure}[tbp]
\centering
\vspace{-2mm}
\includegraphics[width=0.85\linewidth]{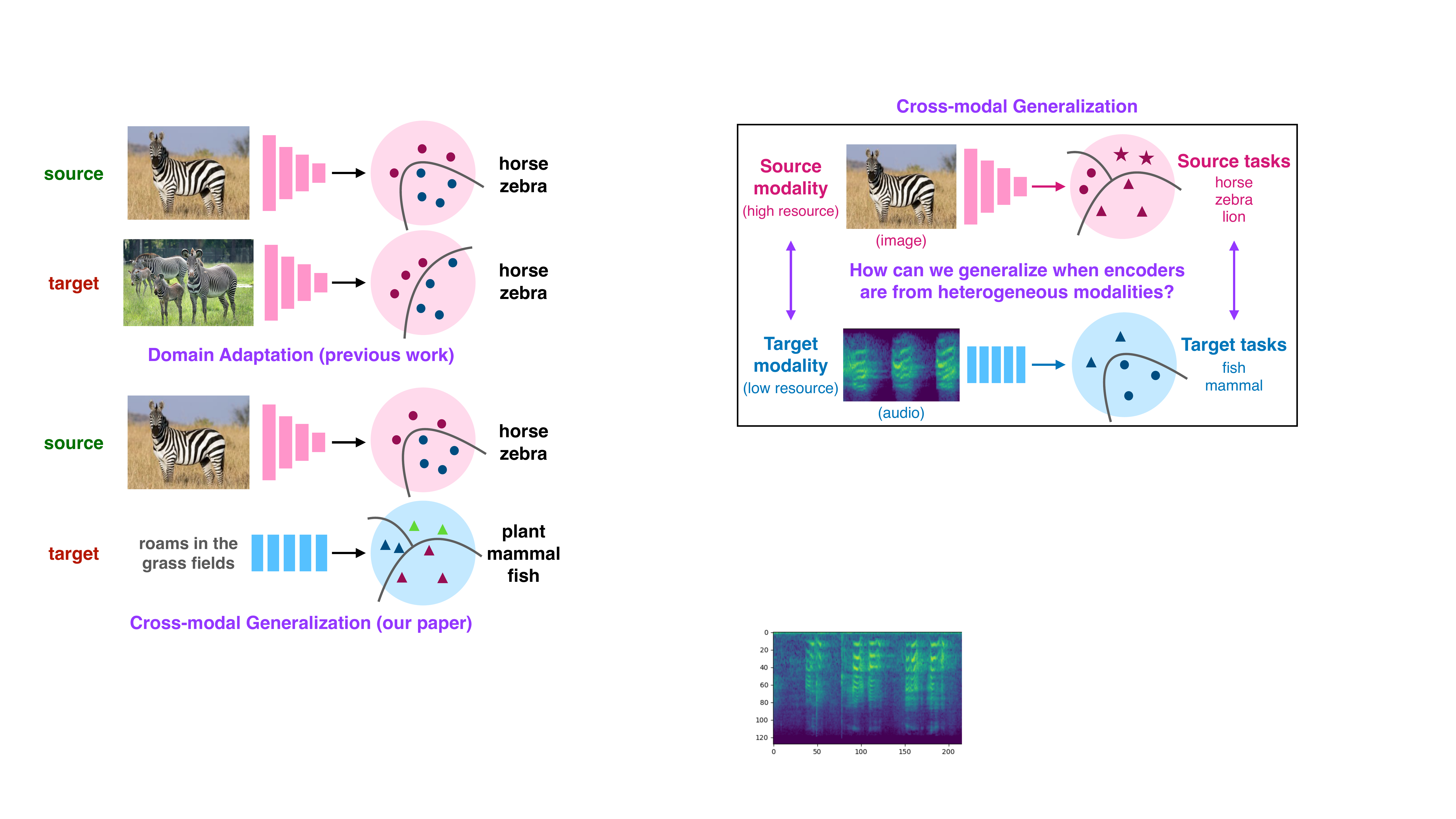}
\vspace{-2mm}
\caption{The \textit{cross-modal generalization} paradigm brings discrepancies in both input and output spaces with \textit{new tasks} expressed in \textit{new modalities}. This raises a fundamental question: how can we ensure generalization across modalities despite using separate encoders for different source (image) and target (audio) modalities? This paper studies the minimal supervision required to perform this alignment and succeed in cross-modal generalization.\vspace{-4mm}}
\label{first_fig}
\end{figure}

\begin{figure}[tbp]
\centering
\vspace{-0mm}
\includegraphics[width=\linewidth]{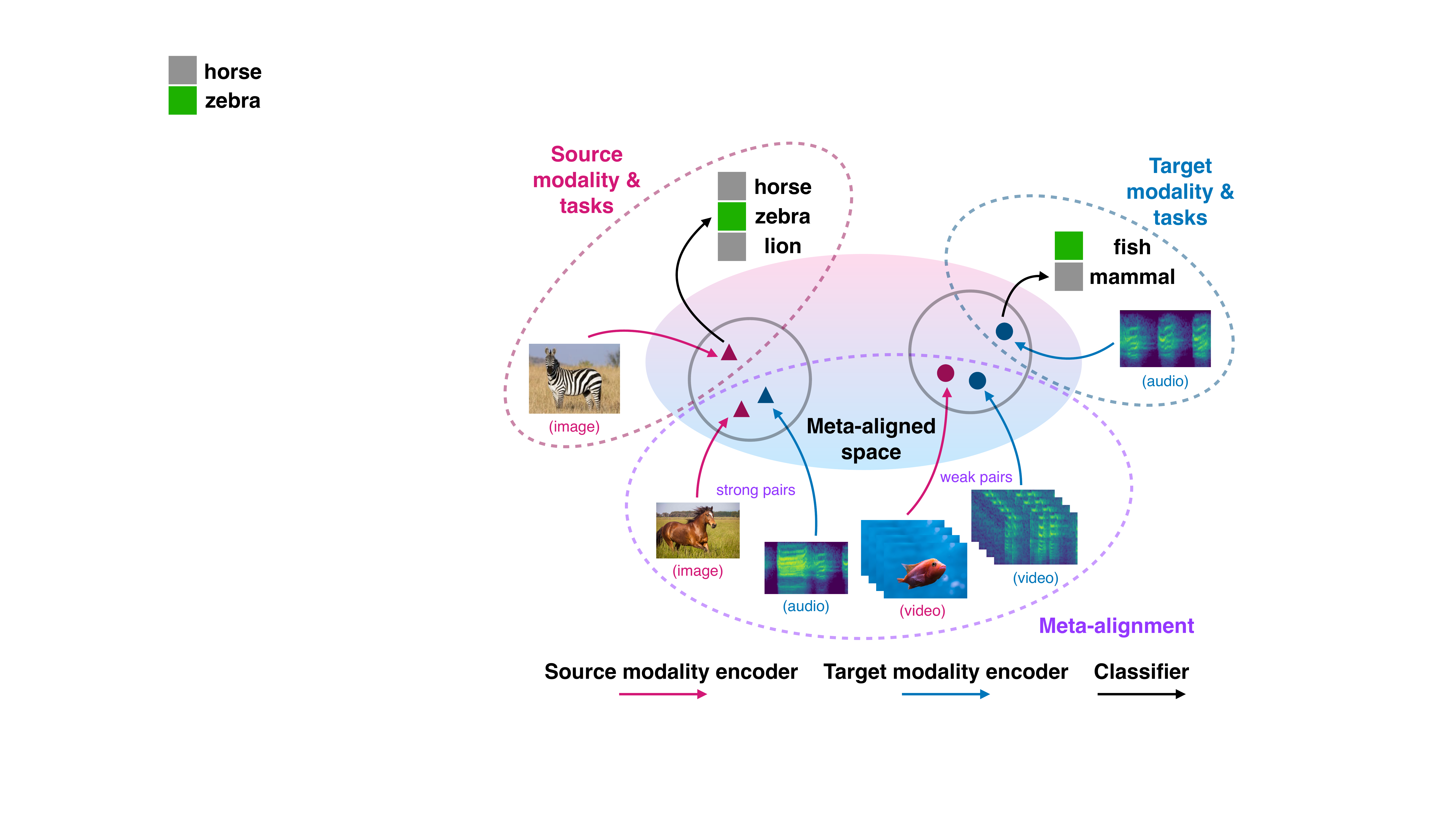}
\vspace{-6mm}
\caption{\textit{Meta-alignment} aligns representation spaces while ensuring quick generalization to new tasks using strongly (image and audio) and weakly (video) paired data. This enables generalization to the target (audio) having only seen labeled data in the source (image), as assessed by few-shot classification and alignment tasks.\vspace{-4mm}}
\label{overview_fig}
\end{figure}

As a motivating example, Figure~\ref{first_fig} illustrates a scenario where large-scale image classification benchmarks can help audio classification, which is a less studied problem with fewer large-scale benchmarks.
In this ambitious problem statement, a key research question becomes: how can we obtain generalization across modalities despite using separate encoders for different source (image) and target (audio) modalities?
The technical challenge involves aligning shared knowledge learned from source image tasks with target audio tasks. Our problem statement differs from conventional meta-learning~\cite{finn2017model} and domain adaptation~\cite{tzeng2017adversarial} where one can take advantage of the same source and target modality with shared encoders which helps generalization by having the same representation space.
In our case, the discrepancies in modalities requires one to learn \textit{new output concepts} expressed in \textit{new input modalities}.
As a result, cross-modal generalization requires new ideas to synchronize (align) multimodal sources and targets.
What is the minimal extra supervision required to perform this alignment?

In this paper, we formalize the conditions required for successful generalization and show that another level of supervision is necessary under partial observability across modalities and tasks. Supervision comes in the form of \textit{cross-modal meta-alignment} (Figure~\ref{overview_fig}) to capture a space where representations of similar concepts in different modalities are close together while ensuring quick generalization to new tasks (i.e. with just a few labels in the target modality).
We introduce a novel algorithm called \names\ (\namel) that leverages readily available multimodal data from the internet (e.g.~\cite{DBLP:journals/corr/KrishnaZGJHKCKL16,perez-rosas_utterance-level_2013,zadeh2018multimodal}) for meta-alignment and cross-modal generalization.
Through theoretical analysis and empirical ablations, we study our proposed algorithm with both strongly and weakly paired multimodal data, showing that cross-modal generalization is possible even with limited extra supervision.

We present experiments on three cross-modal tasks: generalizing from (1) text to image, (2) image to audio, and (3) text to speech.
In all cases, the goal is to classify data from a new target modality given only a few (1-10) labeled samples.
We perform extensive experiments to compare with related approaches including target modality meta-learning that would be expected to perform well since they have seen thousands of labeled examples from the target modality during meta-training.
Surprisingly, \names\ is competitive with these baselines and significantly outperforms other cross-modal approaches.
In addition, we study settings where the target modality suffers from noisy or limited data, a scenario particularly prevalent in low-resource modalities~\cite{DBLP:journals/corr/abs-1802-05368}.

\vspace{-1mm}
\section{Related Work}
\vspace{-1mm}

\definecolor{gg}{RGB}{15,125,15}
\definecolor{rr}{RGB}{190,45,45}

\begin{table}[t]
\fontsize{9}{11}\selectfont
\centering
\vspace{-0mm}
\caption{Data requirements for generalization (i.e. (meta)-test) in a target modality where both data and labels are rare. Cross-modal generalization leverages data from abundant source modalities for low-resource target modalities, requiring only a \textit{few samples} and \textit{no labels} in the target beyond those used for few-shot fine-tuning.}
\vspace{-1em}
\setlength\tabcolsep{2.0pt}
\begin{tabular}{l || c c c}
\Xhline{3\arrayrulewidth}
\multirow{2}{*}{\sc Approaches} & \multicolumn{3}{c}{\sc (Meta-)Train} \\
& Modality & Data & Labels \\
\Xhline{0.5\arrayrulewidth}
Transfer learning~\cite{baevski2019effectiveness} & Target & Many & None\\
\Xhline{0.5\arrayrulewidth}
Unsupervised pre-training~\cite{devlin2018bert} & Target & Many & None\\
Unsupervised meta-learning~\cite{DBLP:journals/corr/abs-1810-02334} & Target & Many & None\\
\Xhline{0.5\arrayrulewidth}
Domain adaptation~\cite{tzeng2017adversarial} & Target & Many & Many\\
Few-shot learning~\cite{finn2017model} & Target & Many & Many\\
\Xhline{0.5\arrayrulewidth}
\multirow{2}{*}{\parbox{4.5cm}{Within modality + cross-modal learning~\cite{DBLP:journals/corr/abs-1903-11101,10.5555/2999611.2999716,DBLP:journals/corr/abs-1710-08347,NIPS2019_8731,zadeh2020foundations}}} & Source & Many & None \\
& Target & Many & Many \\
\Xhline{0.5\arrayrulewidth}
\multirow{2}{*}{\textbf{Cross-modal generalization (ours)}} & Source & Many & Many \\
& \textbf{Target} & \textbf{Few} & \textbf{None} \\
\Xhline{3\arrayrulewidth}
\end{tabular}
\label{overview_table}
\vspace{-4mm}
\end{table}

\textbf{Few-shot learning} has enabled strong performance for settings with limited labeled data~\cite{chen2019closer,DBLP:journals/corr/abs-1806-00388,DBLP:journals/corr/abs-1802-05368} using techniques spanning data augmentation~\cite{antoniou2017data}, metric learning~\cite{snell2017prototypical,vinyals2016matching}, and learning better initializations~\cite{munkhdalai2017meta,ravi2016optimization}. In the latter, \textbf{meta-learning} has recently emerged as a popular choice due to its simplicity in combination with gradient-based methods~\cite{finn2017model}.

\textbf{Transfer learning} focuses on transferring knowledge from external data (e.g. larger datasets~\cite{DBLP:journals/corr/HuhAE16}, unlabeled data~\cite{devlin2018bert}, and knowledge bases~\cite{k-m-etal-2018-learning}) to downstream tasks where labeled data is expensive~\cite{DBLP:journals/corr/abs-1808-01974}. \textbf{Domain adaptation} similarly focuses on changing data distributions~\cite{dong2018domain,hsu2017learning}. However, existing works focus on data within the same modality (i.e. image domain adaptation~\cite{tzeng2017adversarial}, language transfer learning~\cite{devlin2018bert}) which simplifies the alignment problem.

\textbf{Cross-modal alignment} involves learning a joint space where the representations of the same concepts expressed in different modalities are close together~\cite{baltruvsaitis2018multimodal}. Alignment is particularly useful for \textbf{cross-modal retrieval} (e.g. retrieving captions from images)~\cite{frome2013devise} and cross-modal (or cross-lingual) representation learning~\cite{DBLP:journals/corr/abs-1902-09492,wang2019cross}. Several objective functions for learning aligned spaces from varying quantities of paired~\cite{cao2017transitive,faghri2017vse++,huang2017cross} and unpaired~\cite{DBLP:journals/corr/abs-1805-11222} data have been proposed. However, cross-modal generalization is harder since: (1) one has to learn not just the associations between modalities but also associations to labels, (2) there is weak supervision both the target modality and in the label space (see Table~\ref{overview_table}), (3) tasks in different modalities have different (but related) label spaces, and (4) new tasks in the target modality have to be learned using only a few samples.

\textbf{Cross-modal learning:} Recent work has explored more general models that enable knowledge transfer across modalities. In particular, cross-modal data programming~\cite{DBLP:journals/corr/abs-1903-11101} uses weak labels in a source modality to train a classifier in the target modality. Cross-modal transfer learning aims to classify the same task from different input modalities~\cite{huang2017cross,10.1007/978-3-319-68783-4_35}. Finally, few-shot learning within target modalities (e.g. images) has been shown to benefit from additional multimodal information (e.g. word embeddings~\cite{10.5555/2999611.2999716,DBLP:journals/corr/abs-1710-08347,NIPS2019_8731} or videos~\cite{zadeh2020foundations}) during training. However, these all require labeled data from the target modality during meta-training (from a different domain). In contrast, we study \textit{cross-modal} generalization which do not assume \textit{any} labeled data in the target except during few-shot classification.

\vspace{-1mm}
\section{Formalizing Cross-modal Generalization}
\label{formalize}
\vspace{-1mm}

Cross-modal generalization is a learning paradigm to quickly perform new tasks in a target modality despite being trained on a different source modality. To formalize this paradigm, we build on the definition of meta-learning~\cite{hospedales2020meta} and generalize it to study multiple input modalities.
Meta-learning uses labeled data for existing source tasks to enable fast learning on new target tasks~\cite{DBLP:journals/corr/abs-1902-10644}.
We start by defining $M$ different heterogeneous input spaces (modalities) and $N$ different label spaces (tasks).
We denote a modality by an index $m\in \{1,...,M\}$ and a task by $n\in \{1,...,N\}$.

\begin{figure}[tbp]
\centering
\vspace{-2mm}
\includegraphics[width=0.7\linewidth]{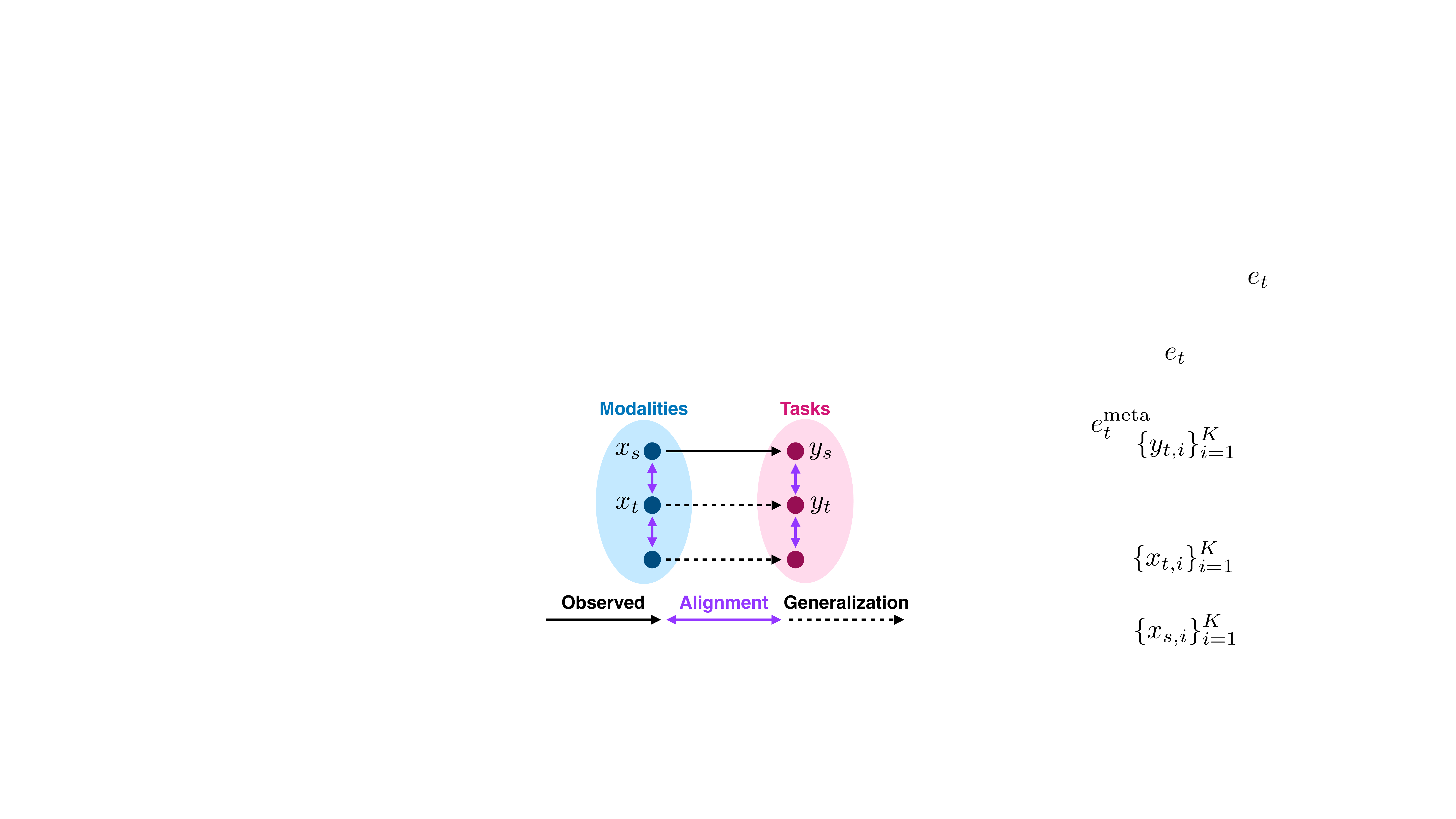}
\vspace{-1em}
\caption{One obtains a subset of observed edges through labeled datasets for source modalities $x_s$ and tasks $y_s$ (solid edges). Generalizing to the target modalities $x_t$ and tasks $y_t$ (dotted edge) requires bridging modalities and tasks through alignment (purple).\vspace{-1em}}
\label{graph}
\end{figure}

Each classification problem $\mathcal{T}(m,n)$ is defined as a triplet with a modality, task, plus a joint distribution: $\mathcal{T}(m,n) = (\mathcal{X}_m, \mathcal{Y}_n, p_{m,n}(x, y))$.
$\mathcal{X}_{m}$ denotes the input space and $\mathcal{Y}_{n}$ the label space sampled from a distribution $p(m,n) := p(\mathcal{X}_{m}, \mathcal{Y}_{n})$ given by a marginal over the entire \textit{meta-distribution}, $p(x_1,..., x_M, y_1, ...y_N, \mathcal{X}_{m_1},...\mathcal{X}_{m_M}, \mathcal{Y}_{n_1},...\mathcal{Y}_{n_N})$. The meta-distribution gives the underlying relationships between all modalities and tasks through a hierarchical generative process $m_i \sim p(m), n_j \sim p(n)$: first picking a modality and task $(m_i,n_j)$ from priors $p(m)$ and $p(n)$ over input and output spaces, before drawing data $x_i$ from $\mathcal{X}_{m_i}$ and labels $y_j$ from $\mathcal{Y}_{n_j}$.
Within each classification problem is also an underlying pairing function mapping inputs to labels through $p_{m,n}(x,y) := p(x,y | m,n)$ for all $x\in \mathcal{X}_{m},\ y\in \mathcal{Y}_{n}$ representing the true data labeling process. To account for generalization over modalities and tasks, cross-modal generalization involves learning a single function $f_w$ with parameters $w$ over the meta-distribution with the following objective:
\begin{definition}
    The \underline{cross-modal generalization problem} is
    \begin{equation}
        \label{eq: meta-learning generalization}
        \argmax_w\mathcal{L}[f_w] := 
        \argmax_w \underset{\substack{m,n\sim p(m,n) \\ x,y\sim p_{m,n}(x,y)}}{\mathbb{E}} \log\left[\frac{f_w(x,y,m,n)}{p(x,y| m,n)}\right].
    \end{equation}
\end{definition}
In practice, the space between modalities and tasks is only \textit{partially observed}: $p(x,y|m,n)$ is only observed for certain modalities and tasks (e.g. labeled classification tasks for images~\cite{deng2009imagenet}, or paired data across image, text, and audio in online videos~\cite{45619}). For other modality-task pairs, we can only obtain inaccurate estimates $q(x,y|m,n)$, often due to having only \textit{limited labeled data}. 
It is helpful to think about this \textit{partial observability} as a bipartite graph $G=(V_x,V_y,E)$ between a modality set $V_x$ and task set $V_y$ (see Figure~\ref{graph}). A solid directed edge from $u \in V_x$ to $v \in V_y$ represents learning a classifier from modality $u$ for task $v$ given an abundance of observed labeled data, which incurs negligible generalization error.
Since it is unlikely for all edges between $V_x$ and $V_y$ to exist, define the \textit{low-resource subset} $\mathcal{M}$ as the complement of $E$ in $V_x \times V_y$. $\mathcal{M}$ represents the set of low-resource modalities and tasks where it is difficult to obtain labeled data.
The focus of cross-modal generalization is to learn a classifier in $\mathcal{M}$ as denoted by a dashed edge. In contrast to solid edges, the lack of data in $\mathcal{M}$ incurs large error along dashed edges.

\begin{figure*}[tbp]
\centering
\vspace{-6mm}
\includegraphics[width=0.9\linewidth]{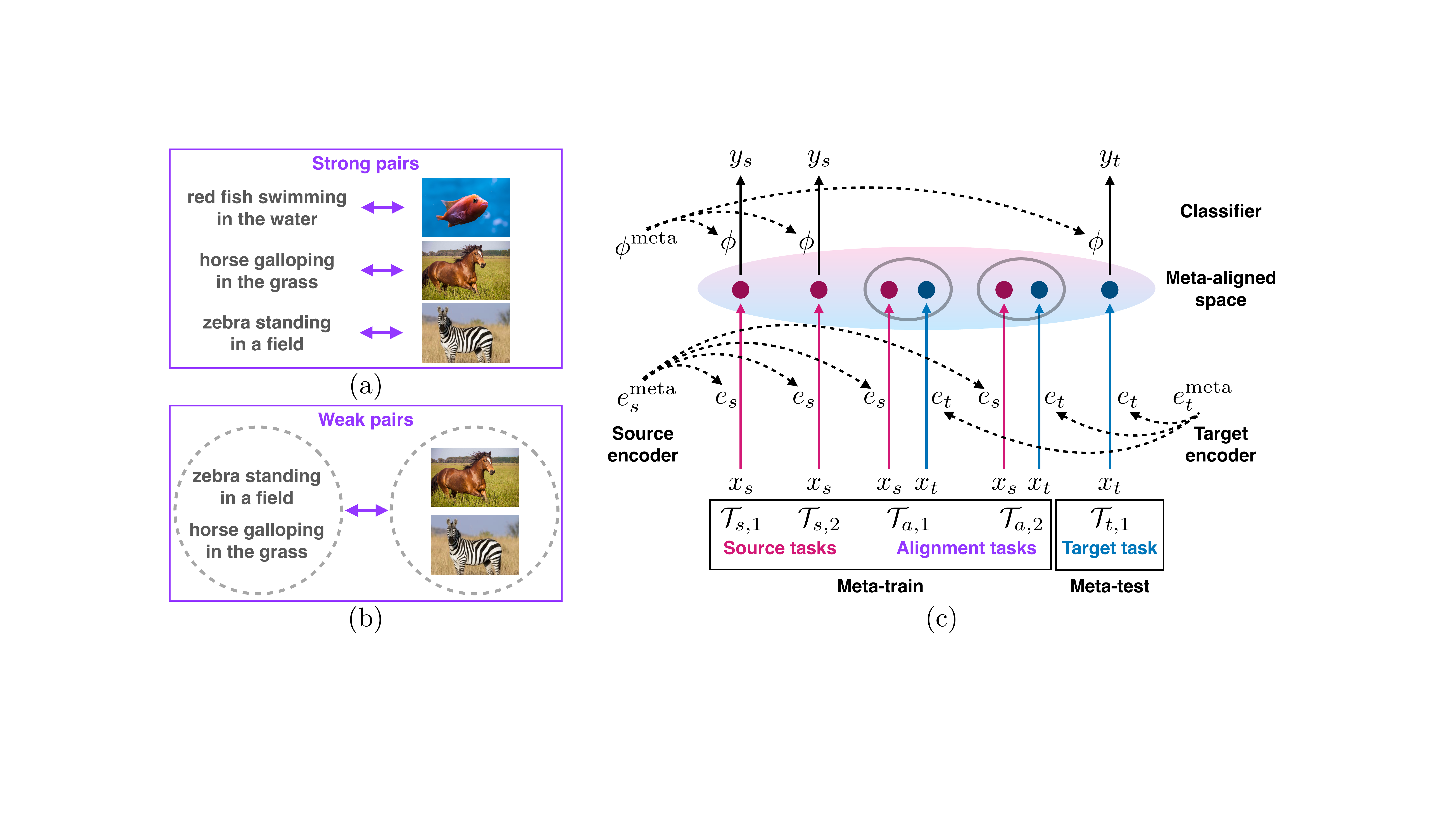}
\vspace{-1em}
\caption{While strong pairs (a) provide exact, one-to-one correspondences across modalities, weak pairs (b) represent coarse semantic groupings which better reflect many-to-many cross-modal mappings and leverage weakly paired multimodal data available on the internet (e.g. videos, image captions). (c) During meta-training, meta-parameters $e_s^\textrm{meta}, e_t^\textrm{meta}, \phi^\textrm{meta}$ are trained using source modality classification tasks $\mathcal{T}_s$ and alignment tasks $\mathcal{T}_a$. Meta-testing uses trained meta-parameters for few-shot generalization to target modality tasks $\mathcal{T}_t$.\vspace{-2mm}}
\label{meta}
\end{figure*}

Therefore, the challenge in cross-modal generalization amounts to finding the path of lowest cumulative error between an input target modality $x_t \in V_x$ and output task $y_t \in V_y$ in $\mathcal{M}$. The key insight is to leverage \textit{cross-modal information} to ``bridge'' modalities that are each labeled for only a subset of tasks (see purple edges in Figure~\ref{graph}). We model cross-modal information as $p(x_s, x_t)$, i.e. \textit{alignment} between modalities $x_s$ and $x_t$, where $x_s$ is a source modality with high-resource data and labels $(x_s, y_s)$. When there is an abundance of paired data $(x_s, x_t)$ (solid purple edge), we say that \textit{strong} alignment exists; otherwise, only \textit{weak} alignment exits. Since strong alignment incurs negligible error in estimating $p(x_s, x_t)$, the alternative \textit{cross-modal path} $P = \{(x_t, x_s), (x_s, y_s), (y_s, y_t)\}$ might link $x_t$ and $y_t$ with \textit{lower} weighted error and is preferable to direct low-resource training for the dashed edge $(x_t, y_t)$. When only weak alignment is available, a trade-off emerges and one has to choose between the error induced by direct low-resource training and the error induced by weak alignment. $(y_s, y_t)$ models relationships across source and target tasks using approaches such as multi-task~\cite{caruana1997multitask} or meta-learning~\cite{finn2017model}. More formally,
\begin{definition}
    Let $p(x_i, x_j)$ be known for $x_i\in \mathcal{D}_{m_i}^{x}$, $x_j\in \mathcal{D}_{m_j}^{x}$ and $i\neq j$. If both $p(x_i|x_j)$ and $p(x_j|x_i)$ are delta distributions (i.e., one-to-one mapping between $x_i$ and $x_j$), there is a \underline{strong alignment} between modality $m_i$ and $m_j$. Otherwise, there is only \underline{weak alignment}.
\end{definition}
We now show that strong alignment can achieve optimal generalization error for tasks in the low-resource set $\mathcal{M}$.
\begin{proposition}
    \label{prop: strong align}
    (Benefit of strong alignment). Let all the modalities be pairwise strongly-aligned, then we can define a surrogate loss function $\Tilde{\mathcal{L}}[f_w]$ such that $\mathcal{L}[\argmax_{w} \Tilde{\mathcal{L}}[f_w] ] = 0$.
\end{proposition}
The proof is provided in Appendix~\ref{formalize_supp}. On the other hand, weak alignment may not provide perfect generalization and we elucidate some conditions when it works in Section~\ref{tradeoff}.

To summarize, cross-modal information $p(x_s, x_t)$ allows us to bridge modalities that are each labeled for only a subset of tasks and achieve generalization to new modalities and tasks in $\mathcal{M}$. In the following section, we explain an algorithm based on contrastive learning~\cite{frome2013devise} to estimate $p(x_s, x_t)$ from data and meta-learning to model $(y_s, y_t)$.








\vspace{-1mm}
\section{\textsc{CroMA}: \namel}
\vspace{-1mm}


Based on our theoretical insights, we propose a practical algorithm for cross-modal generalization involving two thrusts: (1) learning a multimodal space via meta-alignment to model $(x_s,x_t)$ (\S\ref{algo:align}), and (2) learning a cross-modal classifier to model $(y_s, y_t)$ (\S\ref{algo:classify}), which jointly enable generalization to new modalities and tasks. We call our method \names, short for \namel.

\vspace{-1mm}
\subsection{Meta-Alignment}
\label{algo:align}
\vspace{-1mm}

\begin{figure*}
    \begin{minipage}{\linewidth}
    \vspace{-2mm}
    \begin{algorithm}[H]
    \caption{\textsc{CroMA}: \namel}
    \begin{algorithmic}
    \State Initialize meta-alignment encoders $e_s^\textrm{meta}$ and $e_t^\textrm{meta}$, meta-classifier $\phi^\textrm{meta}$.
    \For{iteration = $1,2,\dots$}
        \State Sample alignment task $\mathcal{T}_{a}$ with train $\mathcal{D}^{\mathcal{T}_{a}}_{\textrm{train}}$ and test data $\mathcal{D}^{\mathcal{T}_{a}}_{\textrm{test}}$ of pairs $\{ x_s, x_t \}$.
        \State Initialize $e_s := e_s^\textrm{meta}, e_t := e_t^\textrm{meta}$ and compute alignment loss~\eqref{align_eq} on train data $\mathcal{D}^{\mathcal{T}_{a}}_{\textrm{train}}$.
        \State Compute $\wt{e}_s$ and $\wt{e}_t$ after gradient updates using alignment loss wrt $e_s$ and $e_t$.
        \State Update meta-alignment encoders $e_s^\textrm{meta} \leftarrow e_s^\textrm{meta} + \epsilon (\wt{e}_s - e_s^\textrm{meta})$, $e_t^\textrm{meta} \leftarrow e_t^\textrm{meta} + \epsilon (\wt{e}_t - e_t^\textrm{meta})$.
        \State Sample source modality task $\mathcal{T}_{s}$ with train $\mathcal{D}^{\mathcal{T}_{s}}_{\textrm{train}}$ and test data $\mathcal{D}^{\mathcal{T}_{s}}_{\textrm{test}}$ of pairs $\{ x_s, y_s \}$.
        \State Initialize $\phi := \phi^\textrm{meta}$ and compute classification loss on train data $\mathcal{D}^{\tau_s}_{\textrm{train}}$.
        \State Compute $\wt{\phi}$ after gradient updates using classification loss wrt $\phi$.
        \State Update meta-classifier $\phi^\textrm{meta} \leftarrow \phi^\textrm{meta} + \epsilon (\wt{\phi} - \phi^\textrm{meta})$.
    \EndFor
    \end{algorithmic}
    \label{algo}
    \end{algorithm}
    \end{minipage}
    \vspace{-2mm}
\end{figure*}

We first simplify the problem by assuming access to strong pairs across modalities of the form $(x_s, x_t)$ which makes it easier to learn \textit{strong alignment} (see Figure~\ref{meta}(a)). At the same time, this is not an excessively strong assumption: many multimodal datasets contain paired multimodal data (e.g. activity recognition from audio and video~\cite{45619} and emotion recognition from text, speech, and gestures~\cite{liang2018multimodal,zadeh2018multimodal}).

In practice, we model alignment by learning $p_\theta(x_t | x_s)$. However, directly learning a translation model $p_\theta(x_t | x_s)$ via MLE by mapping each $x_s$ to its corresponding $x_t$ is unlikely to work in practice since $x_s$ and $x_t$ are extremely high-dimensional and heterogeneous data sources which makes reconstruction difficult~\cite{larsen2016autoencoding}. Instead, we use Noise Contrastive Estimation (NCE) which learns a binary classifier to distinguish paired samples $(x_s,x_t) \in \mathcal{D}$ from unpaired negative samples $x_{t, \textrm{neg}}$, which in the asymptotic limit is an unbiased estimator of $p(x_t | x_s)$~\cite{dyer2014notes} but is much easier in practice than generating raw data.

However, the vanilla NCE objective does not handle new tasks at test time. We propose \textit{meta-alignment} to capture an aligned space (i.e. $(x_s, x_t)$) while ensuring quick generalization to new tasks across different modalities (i.e. $(y_s, y_t)$). Meta-alignment trains encoders $e_s, e_t$ for source and target modalities across multiple alignment tasks $\{ \mathcal{T}_{a,1}, ..., \mathcal{T}_{a,T} \}$ into an aligned space. Each alignment task $\mathcal{T}_a$ consists of paired data across source and target modalities. We explicitly train for generalization to new tasks by training meta-alignment parameters $e_s^\textrm{meta}$ and $e_t^\textrm{meta}$ that are used to initialize instances of alignment models for new tasks~\cite{finn2017model}. When presented with a new task, we first initialize task parameters using meta parameters $e_s := e_s^\textrm{meta}$, $e_t := e_t^\textrm{meta}$ before training on the task by optimizing for the NCE loss:
{\fontsize{9.5}{12}\selectfont
\begin{equation}
    \mathcal{L}_\textrm{strong align} = \sum_{(x_s,x_t) \in \mathcal{T}_a} \left( - e_s(x_s)^\top e_t(x_t) + \sum_{x_{t, \textrm{neg}}} e_s(x_s)^\top e_t(x_{t, \textrm{neg}}) \right).
    \label{align_eq}
\end{equation}
}where $x_{t, \textrm{neg}}$ denotes unpaired negative samples. The NCE objective has a nice interpretation as capturing a space where the representations of similar concepts expressed in different modalities are close together, and different concepts in different modalities are far apart~\cite{frome2013devise,DBLP:journals/corr/abs-1806-03560}. The meta-parameters $e_s^\textrm{meta}$ and $e_t^\textrm{meta}$ are updated using first-order gradient information~\cite{nichol2018reptile} so that they gradually become better initializations for new alignment tasks spanning new concepts.

\textbf{Weak pairs:} We now relax the data requirements from strong to \textit{weak pairs}. Instead of one-to-one correspondences, weak pairs represent coarse groupings of semantic correspondence (see Figure~\ref{meta}(b)). This better reflects real-world multimodal data since cross-modal mappings are often many-to-many (e.g. many ways of describing an image, many ways of speaking the same sentence), and are abundant on the internet such as videos constituting weak pairs of image, audio, text~\cite{perez-rosas_utterance-level_2013,zadeh2018multimodal}.
We denote a weak pair as \textit{sets} $X_s$ and $X_t$, and define contrastive loss with an expectation over pairs across the sets, i.e. $x_s, x_t \in X_s \times X_t$:
{\fontsize{9.5}{12}\selectfont
\begin{align}
    &\mathcal{L}_\textrm{weak align} = \\
    &\sum_{\substack{(X_s, X_t)\\ \in \mathcal{T}_a}} \left( - \sum_{\substack{(x_s, x_t)\\ \in X_s \times X_t}} e_s(x_s)^\top e_t(x_t) + \sum_{x_{t, \textrm{neg}}} e_s(x_s)^\top e_t(x_{t, \textrm{neg}}) \right) \nonumber
    \label{align_eq}
\end{align}
}and call this \textit{weak alignment}. We sample several $x_s \in X_s$ and $x_t \in X_t$ to treat as paired samples, and obtain negative pairs $x_{t, \textrm{neg}}$ by sampling outside of the paired sets.

\begin{table*}[t]
\fontsize{9}{11}\selectfont
\centering
\caption{Performance on text to image generalization on Yummly-28K (top), image to audio concept classification from CIFAR to ESC-50 (middle), and text to speech generalization on the Wilderness dataset (bottom). \name\ is on par and sometimes outperforms the oracle target modality meta-learning approach that has seen thousands of labeled target samples, and also outperforms existing unimodal, domain adaptation, and cross-modal baselines. \#Target (labels) denotes the number of target modality samples and labels used during meta-training.}
\vspace{-1em}
\setlength\tabcolsep{2.0pt}
\begin{tabular}{C{1.7cm} L{1.9cm} L{5.2cm} || *{3}{C{1.6cm}} C{2.7cm}}
\Xhline{3\arrayrulewidth}
{\sc Task} & {\sc Type} & {\sc Approach}  & {\sc 1-Shot} & {\sc 5-Shot} & {\sc 10-Shot} & {\sc \#Target (labels)}\\
\Xhline{0.5\arrayrulewidth}
\multirow{6}{*}{\shortstack{{\color{gg}Text}\\(Yummly)\\$\downarrow$\\{\color{rr}Image}\\(Yummly)}} & \multirow{2}{*}{Unimodal} & Pre-training~\cite{baevski2019effectiveness,devlin2018bert} & $33.1 \pm 2.8$ & $36.4 \pm 3.5$ & $49.0 \pm 3.8$ & $0(0)$\\
& & Unsup. meta-learning~\cite{DBLP:journals/corr/abs-1810-02334} (reconstruct) & $37.4 \pm 0.6$ & $41.7 \pm 3.7$ & $49.0 \pm 1.0$ & $5131 (0)$ \\
\cline{2-7}
& \multirow{3}{*}{Cross-modal} & Align + Classify~\cite{cicek2019unsupervised,DBLP:journals/corr/abs-1711-03213,DBLP:journals/corr/RajNT15a,tzeng2017adversarial,10.5555/2283516.2283652} & $37.1 \pm 3.0$ & $40.0 \pm 2.7$ & $47.8 \pm 6.6$ & $5131 (0)$ \\
& & Align + Meta Classify~\cite{sahoo2019metalearning} & $39.4 \pm 2.5$ & $40.0 \pm 2.3$ & $48.8 \pm 7.8$ & $5131 (0)$ \\
& & \textbf{\names\ (ours)} & $\mathbf{39.7 \pm 1.3}$ & $\mathbf{47.1 \pm 3.3}$ & $\mathbf{51.1 \pm 2.1}$ & $5131 (0)$ \\
\cline{2-7}
& \multirow{1}{*}{Oracle} & Within modality generalization~\cite{finn2017model,nichol2018reptile} & $38.9 \pm 2.1$ & $42.1 \pm 1.4$ & $47.9 \pm 5.6$ & $5131 (5131)$ \\
\Xhline{3\arrayrulewidth}
\end{tabular}

\vspace{2mm}

\begin{tabular}{C{1.7cm} L{1.9cm} L{5.2cm} || *{3}{C{1.6cm}} C{2.7cm}}
\Xhline{3\arrayrulewidth}
\multirow{7}{*}{\shortstack{{\color{gg}Image}\\(CIFAR)\\$\downarrow$\\{\color{rr}Audio}\\(ESC-50)}} & \multirow{3}{*}{Unimodal} & Pre-training~\cite{baevski2019effectiveness,devlin2018bert} & $44.2 \pm 0.8$ & $72.3 \pm 0.3$ & $77.4 \pm 1.7$ & $0(0)$\\
& & Unsup. meta-learning~\cite{DBLP:journals/corr/abs-1810-02334} (reconstruct) & $36.3 \pm 1.8$ & $67.3 \pm 0.9$ & $76.6 \pm 2.1$ & $920 (0)$ \\
& & Unsup. meta-learning~\cite{DBLP:journals/corr/abs-1810-02334} (weak labels) & $45.6 \pm 1.3$ & $74.2 \pm 0.3$ & $83.7 \pm 0.1$ & $920 (0)$ \\
\cline{2-7}
& \multirow{3}{*}{Cross-modal} & Align + Classify~\cite{cicek2019unsupervised,DBLP:journals/corr/abs-1711-03213,DBLP:journals/corr/RajNT15a,tzeng2017adversarial,10.5555/2283516.2283652} & $45.3 \pm 0.8$ & $73.9 \pm 2.1$ & $78.8 \pm 0.1$ & $920 (0)$ \\
& & Align + Meta Classify~\cite{sahoo2019metalearning} & $47.2 \pm 0.3$ & $77.1 \pm 0.7$ & $80.4 \pm 0.0$ & $920 (0)$ \\
& & \textbf{\names\ (ours)} & $\mathbf{47.5 \pm 0.2}$ & $\mathbf{85.9 \pm 0.7}$ & $\mathbf{92.7 \pm 0.4}$ & $920 (0)$ \\
\cline{2-7}
& \multirow{1}{*}{Oracle} & Within modality generalization~\cite{finn2017model,nichol2018reptile} & $45.9 \pm 0.2$ & $89.3 \pm 0.4$ & $94.5 \pm 0.3$ & $920 (920)$ \\
\Xhline{3\arrayrulewidth}
\end{tabular}

\vspace{2mm}

\begin{tabular}{C{1.7cm} L{1.9cm} L{5.2cm} || *{3}{C{1.6cm}} C{2.7cm}}
\Xhline{3\arrayrulewidth}
\multirow{10}{*}{\shortstack{{\color{gg}Text}\\(Wilderness)\\$\downarrow$\\{\color{rr}Speech}\\(Wilderness)}} & \multirow{2}{*}{Unimodal} & Pre-training~\cite{baevski2019effectiveness,devlin2018bert} & $55.2 \pm 8.6$ & $73.1 \pm 3.4$ & $84.3 \pm 0.1$ & $0 (0)$\\
& & Unsup. meta-learning~\cite{DBLP:journals/corr/abs-1810-02334} (reconstruct) & $61.5 \pm 4.4$ & $83.5 \pm 4.0$ & $88.5 \pm 2.1$ & $4395 (0)$\\
\cline{2-7}
& \multirow{4}{*}{\shortstack[l]{Domain\\Adaptation}} & Shared~\cite{DBLP:journals/corr/HuhAE16} & $55.6 \pm 10.2$ & $75.2 \pm 8.4$ & $81.9 \pm 3.9$ & $4395 (0)$\\
& & Shared + Align~\cite{jawanpuria2020geometry} & $59.7 \pm 7.6$ & $78.4 \pm 6.2$ & $84.3 \pm 1.5$ & $4395 (0)$\\
& & Shared + Domain confusion~\cite{DBLP:journals/corr/TzengHZSD14} & $59.5 \pm 7.2$ & $76.3 \pm 9.4$ & $83.9 \pm 1.8$ & $4395 (0)$\\
& & Shared + Target labels~\cite{DBLP:journals/corr/abs-0907-1815} & $57.3 \pm 9.3$ & $76.2 \pm 8.4$ & $84.0 \pm 1.9$ & $4395 (4395)$\\
\cline{2-7}
& \multirow{3}{*}{Cross-modal} & Align + Classify~\cite{cicek2019unsupervised,DBLP:journals/corr/abs-1711-03213,DBLP:journals/corr/RajNT15a,tzeng2017adversarial,10.5555/2283516.2283652} & $61.1 \pm 6.0$ & $74.8 \pm 2.1$ & $86.2 \pm 0.7$ & $4395 (0)$\\
& & Align + Meta Classify~\cite{sahoo2019metalearning} & $65.6 \pm 6.1$ & $89.9 \pm 1.5$ & $93.0 \pm 0.5$ & $4395 (0)$\\
& & \textbf{\names\ (ours)} & $\mathbf{67.9 \pm 6.6}$ & $\mathbf{90.6 \pm 1.5}$ & $\mathbf{93.2 \pm 0.2}$ & $4395 (0)$\\
\cline{2-7}
& \multirow{1}{*}{Oracle} & Within modality generalization~\cite{finn2017model,nichol2018reptile} & $61.3 \pm 11.2$ & $77.0 \pm 0.3$ & $87.5 \pm 0.6$ & $4395 (4395)$\\ 
\Xhline{3\arrayrulewidth}
\end{tabular}

\label{tts}
\vspace{0mm}
\end{table*}

\vspace{-1mm}
\subsection{Cross-modal Generalization}
\label{algo:classify}
\vspace{-1mm}

Given a well-aligned space between modalities, we now train a single classifier parametrized by a set of meta-parameters $\phi^\textrm{meta}$ on top of the aligned space for generalization across tasks $(y_s, y_t)$.
The joint set of classification tasks consists of tasks $\{ \mathcal{T}_{s,1}, ..., \mathcal{T}_{s,T} \}$ in the source modality and tasks $\{ \mathcal{T}_{t,1}, ..., \mathcal{T}_{t,T} \}$ in the target.
When presented with a new task, we first initialize the classifier using meta parameters $\phi := \phi^\textrm{meta}$ before training on the task by optimizing for the cross-entropy loss. The meta-parameters $\phi^\textrm{meta}$ are updated using first-order gradient information~\cite{nichol2018reptile} towards better initialization parameters to classify new concepts. Overall, the meta-training stage consists of alignment tasks $\mathcal{T}_a$ and classification tasks in the source modality $\mathcal{T}_s$. The meta-testing stage presents tasks in the target modality $\mathcal{T}_t$. Each task consists of $k$ labeled pairs to simulate an episode of $k$-shot learning. We show the full training algorithm in Algorithm~\ref{algo} and a visual diagram in Figure~\ref{meta}(c).

During testing, a task $\mathcal{T}_t$ is sampled in the target modality. We initialize a new model with the trained meta-alignment encoder $e_t^\textrm{meta}$ and meta-classifier $\phi^\textrm{meta}$, and perform gradient updates with the $k$ labeled samples in the target modality. Note that throughout the entire training process, only $k$ labeled samples in the target modality are presented to \name, which better reflects scarce target modalities where even labeled data for different tasks is difficult to obtain.

\vspace{-1mm}
\section{Experiments}
\vspace{-1mm}


We test generalization from text to image, image to audio, and text to speech classification tasks. Anonymized code is included in the supplementary. Experimental details and additional results are included in Appendix~\ref{details_supp} and ~\ref{results_supp}.

\vspace{-1mm}
\subsection{Datasets and Tasks}
\label{recipe_exp}
\vspace{-1mm}

\textbf{Text to Image Dataset:} We use the Yummly-28K dataset~\cite{yummly_dataset} which contains parallel text descriptions and images of recipes. We create classification labels from the metadata by concatenating the meal type and cuisine, yielding $44$ distinct classes. The large number of recipes and shared concepts between text and image makes it an ideal testbed for cross-modal generalization.
We used a ResNet pretrained on ImageNet~\cite{deng2009imagenet} to encode the images, pretrained BERT encoder~\cite{devlin2018bert} for text, and a shared network for prediction.

\textbf{Image to Audio Dataset:} We combine two large unimodal classification datasets over images (CIFAR-$10$ and CIFAR-$100$~\cite{krizhevsky2009learning}) and audio made by various objects (ESC-$50$~\cite{piczak2015esc}) with partially related label spaces. This allows us to leverage complementary information from both modalities while testing on new concepts. To obtain weak pairs, we map similar classes between the datasets using similarities from WordNet~\cite{Miller:1995:WLD:219717.219748} and text cooccurrence. This yields $17$ unique clusters of weak pairs (Appendix~\ref{image_audio_supp} lists all the clusters). We used a ResNet pretrained on ImageNet~\cite{deng2009imagenet} to encode the images and a convolutional network pretrained on AudioSet~\cite{gemmeke2017audio} to encode audio~\cite{kumar2018knowledge,ridnik2020tresnet}.

\textbf{Text to Speech Dataset:} We use the Wilderness dataset, a large-scale multimodal dataset composed of parallel multilingual text and speech data~\cite{8683536}. We use a subset of $99$ languages for language classification from text (source) and speech (target) individually. The tasks are split such as there is no overlap between the text and speech samples used for classification and the pairs seen for strong alignment. We use LSTMs to encode both text and speech data.

\textbf{Metrics:} We report few-shot ($k=1,5,10$) classification accuracy in the target modality by fixing $8$ evaluation tasks, each comprised of $5$ unseen target concepts during meta-test. We compute accuracy across all $8$ tasks and repeat experiments $10$ times to report mean and standard deviations.

\textbf{Baselines:} We compare with $4$ broad sets of baselines:

1) \textbf{Unimodal} baselines only use unlabeled data from the target modality during meta-training following our low-resource assumption. The simplest baseline ignores meta-training and just fine-tunes on the tasks in meta-test starting from a (supervised~\cite{baevski2019effectiveness} or unsupervised~\cite{devlin2018bert}) \textbf{pre-trained} model. To better leverage unlabeled target modality data, we also compare with \textbf{unsupervised meta-learning}~\cite{DBLP:journals/corr/abs-1810-02334} which performs self-supervised learning via reconstruction or weak labels during meta-training (see Appendix~\ref{image_audio_supp}).

2) We modify \textbf{Domain Adaptation} (DA) methods to verify that it is necessary to use separate encoders and perform explicit alignment: a) \textbf{Shared} shares all encoder layer for both modalities except a separate linear layer that maps data from the target modality's input dimension to the source~\cite{DBLP:journals/corr/HuhAE16,tzeng2017adversarial}. b) \textbf{Shared + Align} further adds our alignment loss (contrastive loss) on top of the encoded representations, in a manner similar to~\cite{jawanpuria2020geometry}. c) \textbf{Shared + Domain confusion} further adds a domain confusion loss on top of the encoded representations~\cite{DBLP:journals/corr/TzengHZSD14}. d) \textbf{Shared + Target labels} also uses target modality labels during meta-training, similar to supervised DA~\cite{DBLP:journals/corr/abs-0907-1815} (details in Appendix~\ref{vs_domain}).

\definecolor{gg}{RGB}{15,125,15}
\definecolor{rr}{RGB}{190,45,45}

\begin{table*}[t]
\fontsize{9}{11}\selectfont
\centering
\caption{Language classification predictions on low-resource speech samples after training on labeled text data. Despite seeing just $5$ labeled speech samples, our method is able to accurately classify low-resource languages.\vspace{-1em}}
\setlength\tabcolsep{4.0pt}
\begin{tabular}{l || c c c c c c}
\Xhline{3\arrayrulewidth}
{\sc Speech (text in parenthesis)} & {\sc Oracle} & {\sc \names\ (ours)}\\
\Xhline{0.5\arrayrulewidth}
(Beda Yesus agot gu ofa oida Bua buroru Didif ojgomu) & {\color{rr} Russian} & {\color{gg} Meax} \\
(Ido hai Timotiu natile hampai moula Aturana Musa) & {\color{rr} Jamaican Patois} & {\color{gg} Badaic} \\
(Mu habotu pa kali Mataoqu osolae vekoi Rau sari Mua kana pa kauru Nenemu gua) & {\color{rr} Avokaya} & {\color{gg} Roviana}\\
\Xhline{3\arrayrulewidth}
\end{tabular}
\label{samples}
\vspace{-4mm}
\end{table*}

3) \textbf{Cross-modal:} We adapt domain adaptation and meta-learning for cross-modal generalization under the following categories: a) \textbf{Align + Classify} which uses supervised alignment methods such as adversarial learning~\cite{tzeng2017adversarial}, cycle reconstruction~\cite{cicek2019unsupervised,DBLP:journals/corr/abs-1711-03213}, or contrastive loss~\cite{10.5555/2283516.2283652} to align input spaces from multiple domains before training a shared classifier~\cite{DBLP:journals/corr/RajNT15a}. b) \textbf{Align + Meta Classify} which learns a shared space using standard supervised alignment~\cite{frome2013devise} before meta-learning a classifier~\cite{sahoo2019metalearning}, and c) \textbf{\name} which represents our full model of jointly training for generalization across alignment and classification tasks. Since all methods are agnostic to the specific alignment algorithm used, we use contrastive loss with negative sampling as described in Section~\ref{algo:align} for fair comparison across all baselines.

4) \textbf{Oracle:} The ideal (but likely unrealistic) scenario where meta-training and meta-testing both have labeled data in the target modality. We use the Reptile algorithm~\cite{nichol2018reptile} for target modality meta-learning. Since there is the least domain shift, we expect this method to perform best but with the requirement of large amounts of labeled target data.

\begin{figure}[tbp]
\centering
\vspace{-0mm}
\includegraphics[width=\linewidth]{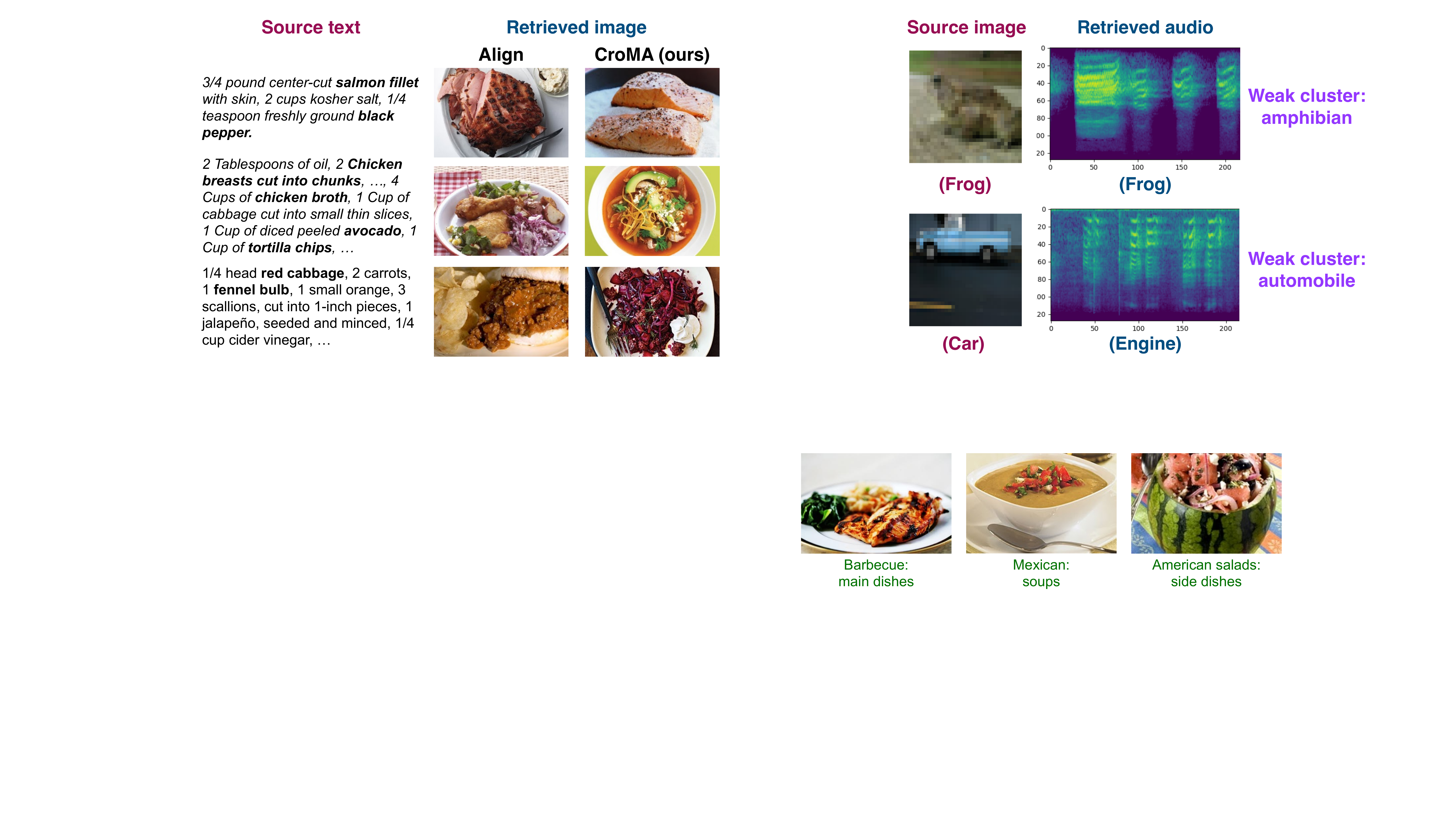}
\vspace{-6mm}
\caption{On Yummly-28K dataset, \names\ leverages source text modality to make accurate few-shot predictions on target image modality despite only seeing $1-10$ labeled image examples.\vspace{-4mm}}
\label{examples}
\end{figure}

\vspace{-1mm}
\subsection{Cross-modal Generalization}
\vspace{-1mm}

\textbf{Comparison to oracle:} For text to image (Table~\ref{tts} top) and text to speech (Table~\ref{tts} bottom), \name\ surprisingly outperforms even the oracle baseline, in addition to unimodal and cross-modal methods. We hypothesize this is because text data (source) is cleaner than image and speech data (target) and the community has better models for encoding text than images and speech spectrograms. Consistent with this hypothesis, we found that text classifiers performed better on Yummly-28K and Wilderness datasets than image and speech classifiers. This implies that \textbf{one can leverage abundant, cleaner, and more-predictive source modalities to improve target modality performance}.
For image to audio (Table~\ref{tts} middle), we observe that our cross-modal approach is on par (outperforms for $k=1$, and within $2-3\%$ for $k=5,10$) with the oracle baseline that has seen a thousand labeled audio examples during meta-training.

\textbf{Comparison to existing approaches:} For all setups, \names\ consistently outperforms existing unimodal and cross-modal baselines. Since we use the same LSTM architecture for both text and speech, we can also apply DA approaches which share encoders. From Table~\ref{tts} bottom, we see that they do not perform well on cross-modal generalization. Although domain confusion and alignment improve upon standard encoder sharing, they still fall short of our approach. Our method also outperforms the Shared + Target labels baseline which further uses target modality labels to train the shared encoder during meta-training. This serves to highlight the important differences between cross-modal generalization and domain adaptation: 1. \textbf{separate encoders} and 2. \textbf{explicit alignment are important}.

\textbf{Ablation studies:} Consistent across all setups in Table~\ref{tts}, we find that jointly meta-training across alignment and classification improves upon standard supervised alignment methods commonly used in domain adaptation~\cite{DBLP:journals/corr/RajNT15a,sahoo2019metalearning}. We find that performance improvement is greatest for the $1$-shot setting, suggesting that meta-alignment is particularly suitable for low-resource target modalities.

\begin{figure*}[tbp]
\centering
\includegraphics[width=0.9\linewidth]{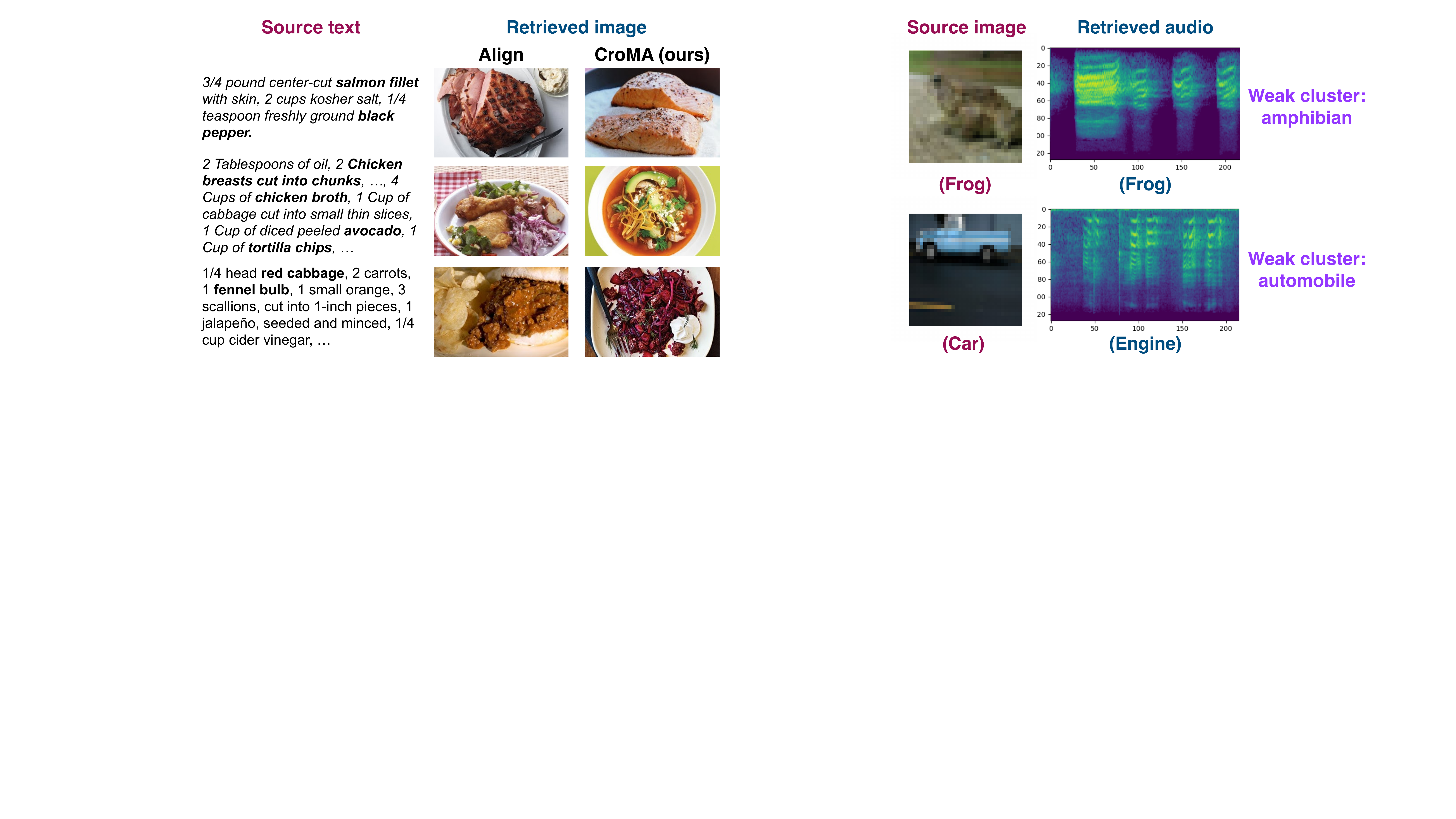}
\vspace{-2mm}
\caption{\textbf{Left}: samples of retrieved images given text recipes. \names\ performs few-shot retrieval of images more accurately than existing alignment approaches. \textbf{Right}: samples of retrieved audio samples given images. Despite being trained only on \textit{weak pairs}, meta-alignment can perform few-shot cross-modal retrieval at fine granularities (e.g. amphibian, automobile).\vspace{-4mm}}
\label{pairs}
\end{figure*}

\definecolor{gg}{RGB}{15,125,15}
\definecolor{rr}{RGB}{190,45,45}

\begin{table}[t!]
\vspace{-0mm}
\fontsize{9}{11}\selectfont
\caption{\name\ yields better alignment scores than the baselines, indicating that meta-alignment can align new concepts using only \textit{weakly paired data} across image and audio.}
\vspace{-1em}
\setlength\tabcolsep{3.0pt}
\begin{tabular}{l l || c c c c c}
\Xhline{3\arrayrulewidth}
{\sc $K$} & {\sc Experiment} & {\sc R$@1$} \textcolor{gg}{$\uparrow$} & {\sc R$@5$} \textcolor{gg}{$\uparrow$} & {\sc R$@10$} \textcolor{gg}{$\uparrow$} & {\sc Rank} \textcolor{gg}{$\downarrow$} & {\sc Cos.} \textcolor{gg}{$\downarrow$} \\
\Xhline{0.5\arrayrulewidth} 
\multirow{3}{*}{$5$} & No align & $1.0\%$ & $2.0\%$ & $5.5\%$ & $101$ & $0.428$\\
& Align & $2.0\%$ & $5.5\%$ & $8.5\%$ & $103$ & $0.272$\\
& \textbf{\names} & $\mathbf{4.0\%}$ & $\mathbf{19.5\%}$ & $\mathbf{39.0\%}$ & $\mathbf{13}$ & $\mathbf{0.003}$\\
\Xhline{0.5\arrayrulewidth}
\multirow{3}{*}{$10$} & No align & $0.5\%$ & $3.0\%$ & $4.5\%$ & $101$ & $0.399$\\
& Align & $1.5\%$ & $11.0\%$ & $18.5\%$ & $52$ & $0.222$\\
& \textbf{\names} & $\mathbf{3.5\%}$ & $\mathbf{17.5\%}$ & $\mathbf{35.0\%}$ & $\mathbf{15}$ & $\mathbf{0.004}$\\
\Xhline{3\arrayrulewidth}
\end{tabular}
\label{image_audio_align}
\vspace{-4mm}
\end{table}

\textbf{Model predictions:} We show some samples of language classification predictions on low-resource speech samples in Table~\ref{samples}. Despite seeing just $5$ labeled speech samples, our method is able to quickly generalize and classify low-resource languages. On the text to image task (Figure~\ref{examples}), \names\ also quickly recognizes images from new recipes.

\begin{figure}[tbp]
\centering
\includegraphics[width=0.7\linewidth]{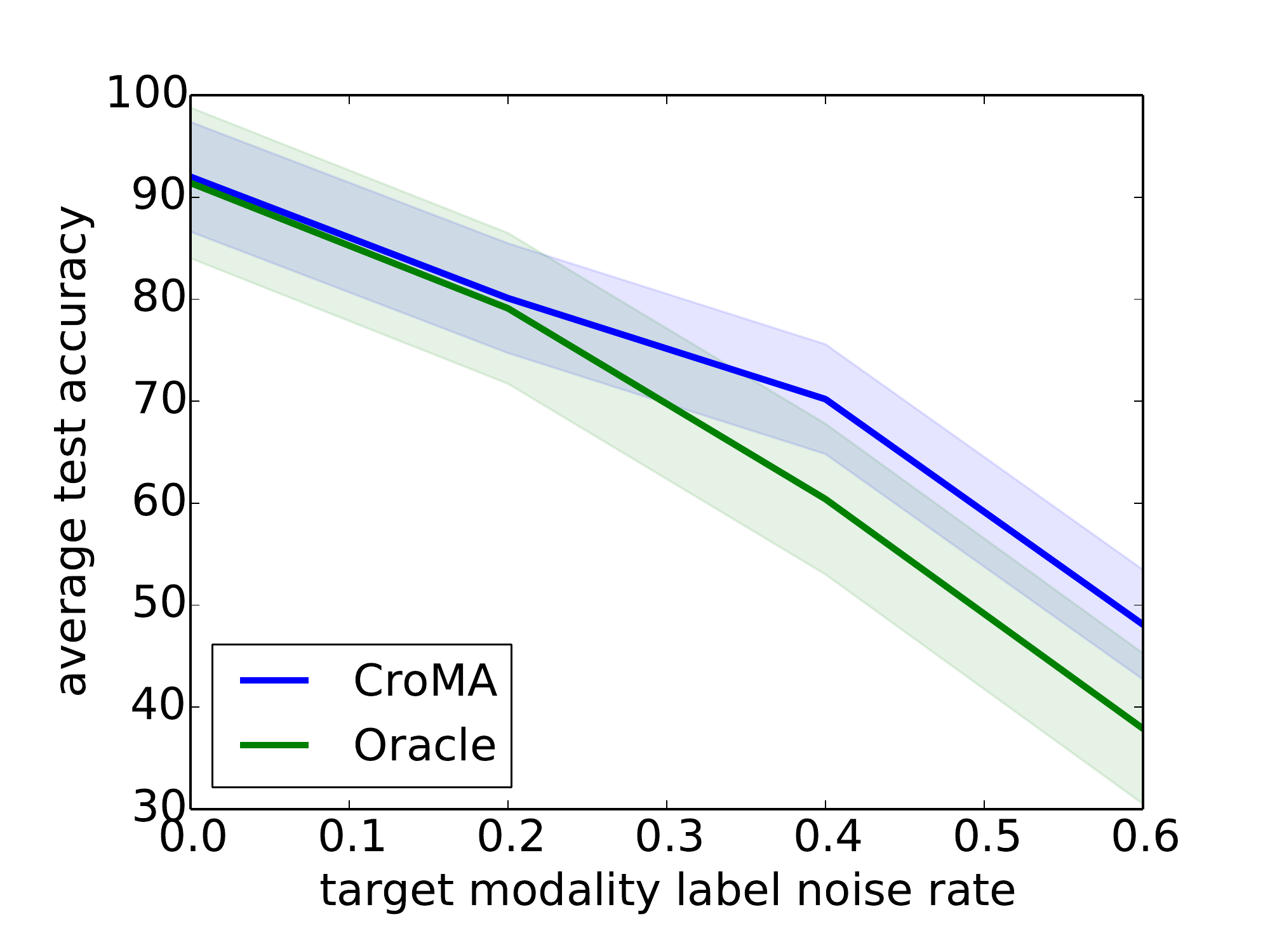}
\vspace{-2mm}
\caption{\name\ is robust to noisy labels in the target modality by using cross-modal information from the source, making it suitable for low-resource modalities with imperfect annotations.\vspace{-4mm}}
\label{noise}
\end{figure}

\vspace{-1mm}
\subsection{Few-shot Cross-modal Retrieval}
\vspace{-1mm}

We show retrieval performance in Table~\ref{image_audio_align} for both standard and meta-alignment strategies as measured using recall$@k$, rank, and cosine loss metrics~\cite{frome2013devise}. Our model yields better retrieval performance than the baselines, indicating that meta-alignment successfully aligns new concepts in low-resource target modalities. In Figure~\ref{pairs}, we also show samples of retrieved data in the target given input in the source modality to help us understand which source modalities the model is basing its target predictions on. Despite being trained only on weak pairs, \textbf{meta-alignment is able to perform cross-modal retrieval at fine granularities}.

\vspace{-1mm}
\subsection{Noisy Target Labels}
\vspace{-1mm}

We also evaluate the effect of noisy labels in the target modality since it is often difficult to obtain exact labels in low-resource modalities such as rare languages. To simulate label noise, we add symmetric noise~\cite{DBLP:journals/corr/abs-1804-06872} to all target modality labels (both meta-train and meta-test). Despite only seeing $k=1,5,10$ labels in the target, \name\ is \textbf{more robust to noisy label} than the oracle baseline (see Figure~\ref{noise}).

\vspace{-1mm}
\subsection{On Alignment vs Supervision}
\label{tradeoff}
\vspace{-1mm}

Finally, to study the tradeoffs in cross-modal alignment, we perform a controlled experiment on synthetic data from 2 modalities: source $D_{1}^{\rm sup} = \{(x_1^i, y_1^i) \}_{i=1}^{n_1}$ and target $D_{2}^{\rm sup} = \{(x_2^i, y_2^i) \}_{i=1}^{n_2}$. The labels are generated via a noisy teacher model $y_m^i = u_m x_m^i + \epsilon_m^i$, where $x_m^i \in \mathbb{R}^d$, $u_m \in \mathbb{R}^d$, and $\epsilon_m^i \sim \mathcal{N}(0, \sigma^2)$ for $m \in \{1,2\}$~\cite{liang2020think}. We model cross-modal and task relationships through a full-rank transformation $x_1^i = W x_2^i$ and $u_1 x_1^i = u_2 W x_2^i$ respectively.

Suppose $n_1 \gg n_2$ (i.e. high-resource source task, low-resource target task). One can train separate supervised models $f_m(x) = w_m x$ and measure the total generalization loss $L = \sum_{m=1}^2 \mathbb{E}_{x_m}\left[(f_m(x_m) - u_m x_m)^2 \right]$, but this loss will be very high in the low-resource target task. Instead, cross-modal alignment learns the transformation $W$ using pairs $D_{\rm unsup}= \{(x_1^i, x_2^i)\}_{i=1}^{n_{\rm align}}$ generated via $x_1^i = W x_2^i + \eta^i$ with noise $\eta^i \sim \mathcal{N}(0, \sigma_W^2)$. $\eta^i$ models uncertainty in alignment pairs: $\sigma_W^2 \rightarrow 0$ represents strong alignment and large $\sigma_W^2$ represents weak alignment.

\begin{figure}[tbp]
\centering
\includegraphics[width=0.75\linewidth]{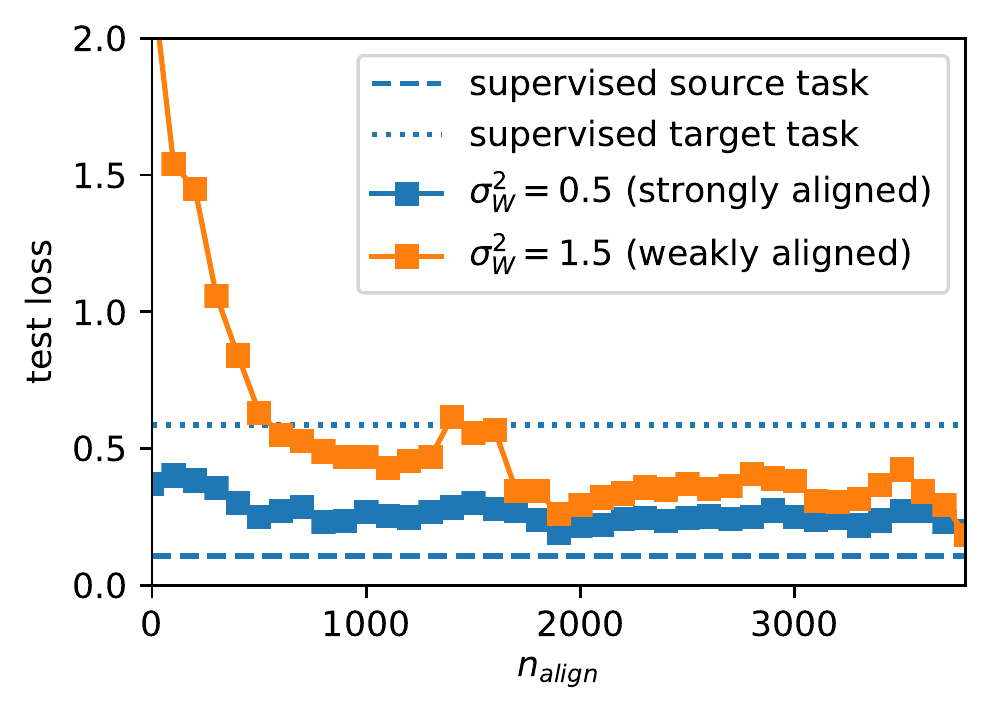}
\vspace{-2mm}
\caption{Supervision learning vs alignment for synthetic data: with more strongly or weakly aligned pairs ($n_{\rm align}$), cross-modal alignment improves upon supervised learning in target tasks.\vspace{-4mm}}
\label{syn}
\end{figure}

We empirically study this setup in Figure~\ref{syn}, where we set $d=20, n_1=250, n_2=40$ and vary $n_{\rm align}$. We observe that (1) more alignment pairs help, but at most by the performance of the high-resource source task, (2) quality of alignment matters: less noise $\sigma_W^2$ in alignment data gives better performance, and (3) even weak alignment is preferable to supervised learning with enough weakly paired data. In fact, under this simplified setup, an analysis shows that training on the high-resource task has error ${d\sigma^2}/{n_1}$ while the low-resource task has error ${d\sigma^2}/{n_2}$. Estimating the alignment matrix with $d^2$ elements results in error ${d^2\sigma^2_W}/{n_{\rm align}}$. Therefore, cross-modal alignment has error $\frac{d^2\sigma^2_W}{n_{\rm align}} + \frac{d\sigma^2}{n_1}$, which should be preferred when $\frac{d\sigma^2_W}{n_{\rm align}} + \frac{\sigma^2}{n_1} < \frac{\sigma^2}{n_2}$. This gives a simple rule-of-thumb for practitioners to choose between supervised learning and cross-modal learning (see Appendix~\ref{tradeoff_supp} for more details and experiments).


\vspace{-1mm}
\section{Conclusion}
\vspace{-1mm}

In this work, we proposed \textit{cross-modal generalization}: a learning paradigm where abundant source modalities are used to help low-resource target modalities. We showed that \textit{meta-alignment} using cross-modal data can allow quick generalization to new concepts across different modalities. Our experiments demonstrate strong performance on classifying data from an entirely new target modality under limited samples and noisy labels, which is particularly useful for generalization to low-resource images, speech, and languages.

\bibliographystyle{ieee_fullname}
\bibliography{refs}

\begin{thebibliography}{10}\itemsep=-1pt

\bibitem{45619}
Sami Abu-El-Haija, Nisarg Kothari, Joonseok Lee, Apostol~(Paul) Natsev, George
  Toderici, Balakrishnan Varadarajan, and Sudheendra Vijayanarasimhan.
\newblock Youtube-8m: A large-scale video classification benchmark.
\newblock In {\em arXiv:1609.08675}, 2016.

\bibitem{antoniou2017data}
Antreas Antoniou, Amos Storkey, and Harrison Edwards.
\newblock Data augmentation generative adversarial networks.
\newblock {\em arXiv preprint arXiv:1711.04340}, 2017.

\bibitem{baevski2019effectiveness}
Alexei Baevski, Michael Auli, and Abdelrahman Mohamed.
\newblock Effectiveness of self-supervised pre-training for speech recognition.
\newblock {\em arXiv preprint arXiv:1911.03912}, 2019.

\bibitem{baltruvsaitis2018multimodal}
Tadas Baltru{\v{s}}aitis, Chaitanya Ahuja, and Louis-Philippe Morency.
\newblock Multimodal machine learning: A survey and taxonomy.
\newblock {\em IEEE transactions on pattern analysis and machine intelligence},
  41(2):423--443, 2018.

\bibitem{8683536}
A.~W. {Black}.
\newblock Cmu wilderness multilingual speech dataset.
\newblock In {\em ICASSP 2019 - 2019 IEEE International Conference on
  Acoustics, Speech and Signal Processing (ICASSP)}, pages 5971--5975, 2019.

\bibitem{cao2017transitive}
Zhangjie Cao, Mingsheng Long, Jianmin Wang, and Qiang Yang.
\newblock Transitive hashing network for heterogeneous multimedia retrieval.
\newblock In {\em Proceedings of the Thirty-First AAAI Conference on Artificial
  Intelligence}, pages 81--87, 2017.

\bibitem{caruana1997multitask}
Rich Caruana.
\newblock Multitask learning.
\newblock {\em Machine learning}, 28(1):41--75, 1997.

\bibitem{chen2019closer}
Wei-Yu Chen, Yen-Cheng Liu, Zsolt Kira, Yu-Chiang~Frank Wang, and Jia-Bin
  Huang.
\newblock A closer look at few-shot classification.
\newblock {\em arXiv preprint arXiv:1904.04232}, 2019.

\bibitem{DBLP:journals/corr/abs-1911-01547}
Fran{\c{c}}ois Chollet.
\newblock On the measure of intelligence.
\newblock {\em CoRR}, abs/1911.01547, 2019.

\bibitem{cicek2019unsupervised}
Safa Cicek and Stefano Soatto.
\newblock Unsupervised domain adaptation via regularized conditional alignment.
\newblock In {\em Proceedings of the IEEE International Conference on Computer
  Vision}, pages 1416--1425, 2019.

\bibitem{deng2009imagenet}
Jia Deng, Wei Dong, Richard Socher, Li-Jia Li, Kai Li, and Li Fei-Fei.
\newblock Imagenet: A large-scale hierarchical image database.
\newblock In {\em 2009 IEEE conference on computer vision and pattern
  recognition}, pages 248--255. Ieee, 2009.

\bibitem{devlin2018bert}
Jacob Devlin, Ming-Wei Chang, Kenton Lee, and Kristina Toutanova.
\newblock Bert: Pre-training of deep bidirectional transformers for language
  understanding.
\newblock {\em arXiv preprint arXiv:1810.04805}, 2018.

\bibitem{dong2018domain}
Nanqing Dong and Eric~P Xing.
\newblock Domain adaption in one-shot learning.
\newblock In {\em Joint European Conference on Machine Learning and Knowledge
  Discovery in Databases}, pages 573--588. Springer, 2018.

\bibitem{DBLP:journals/corr/abs-1903-11101}
Jared Dunnmon, Alexander Ratner, Nishith Khandwala, Khaled Saab, Matthew
  Markert, Hersh Sagreiya, Roger~E. Goldman, Christopher Lee{-}Messer,
  Matthew~P. Lungren, Daniel~L. Rubin, and Christopher R{\'{e}}.
\newblock Cross-modal data programming enables rapid medical machine learning.
\newblock {\em CoRR}, abs/1903.11101, 2019.

\bibitem{dyer2014notes}
Chris Dyer.
\newblock Notes on noise contrastive estimation and negative sampling.
\newblock {\em arXiv preprint arXiv:1410.8251}, 2014.

\bibitem{faghri2017vse++}
Fartash Faghri, David~J Fleet, Jamie~Ryan Kiros, and Sanja Fidler.
\newblock Vse++: Improving visual-semantic embeddings with hard negatives.
\newblock {\em arXiv preprint arXiv:1707.05612}, 2017.

\bibitem{finn2017model}
Chelsea Finn, Pieter Abbeel, and Sergey Levine.
\newblock Model-agnostic meta-learning for fast adaptation of deep networks.
\newblock In {\em Proceedings of the 34th International Conference on Machine
  Learning-Volume 70}, pages 1126--1135. JMLR. org, 2017.

\bibitem{frome2013devise}
Andrea Frome, Greg~S Corrado, Jon Shlens, Samy Bengio, Jeff Dean, Marc'Aurelio
  Ranzato, and Tomas Mikolov.
\newblock Devise: A deep visual-semantic embedding model.
\newblock In {\em Advances in neural information processing systems}, pages
  2121--2129, 2013.

\bibitem{gemmeke2017audio}
Jort~F Gemmeke, Daniel~PW Ellis, Dylan Freedman, Aren Jansen, Wade Lawrence,
  R~Channing Moore, Manoj Plakal, and Marvin Ritter.
\newblock Audio set: An ontology and human-labeled dataset for audio events.
\newblock In {\em 2017 IEEE International Conference on Acoustics, Speech and
  Signal Processing (ICASSP)}, pages 776--780. IEEE, 2017.

\bibitem{DBLP:journals/corr/abs-1806-00388}
Marzyeh Ghassemi, Tristan Naumann, Peter Schulam, Andrew~L. Beam, and Rajesh
  Ranganath.
\newblock Opportunities in machine learning for healthcare.
\newblock {\em CoRR}, abs/1806.00388, 2018.

\bibitem{DBLP:journals/corr/abs-1805-11222}
Edouard Grave, Armand Joulin, and Quentin Berthet.
\newblock Unsupervised alignment of embeddings with wasserstein procrustes.
\newblock {\em CoRR}, abs/1805.11222, 2018.

\bibitem{DBLP:journals/corr/abs-1802-05368}
Jiatao Gu, Hany Hassan, Jacob Devlin, and Victor O.~K. Li.
\newblock Universal neural machine translation for extremely low resource
  languages.
\newblock {\em CoRR}, abs/1802.05368, 2018.

\bibitem{DBLP:journals/corr/abs-1804-06872}
Bo Han, Quanming Yao, Xingrui Yu, Gang Niu, Miao Xu, Weihua Hu, Ivor~W. Tsang,
  and Masashi Sugiyama.
\newblock Co-sampling: Training robust networks for extremely noisy
  supervision.
\newblock {\em CoRR}, abs/1804.06872, 2018.

\bibitem{DBLP:journals/corr/abs-1711-03213}
Judy Hoffman, Eric Tzeng, Taesung Park, Jun{-}Yan Zhu, Phillip Isola, Kate
  Saenko, Alexei~A. Efros, and Trevor Darrell.
\newblock Cycada: Cycle-consistent adversarial domain adaptation.
\newblock {\em CoRR}, abs/1711.03213, 2017.

\bibitem{hospedales2020meta}
Timothy Hospedales, Antreas Antoniou, Paul Micaelli, and Amos Storkey.
\newblock Meta-learning in neural networks: A survey.
\newblock {\em arXiv preprint arXiv:2004.05439}, 2020.

\bibitem{DBLP:journals/corr/abs-1810-02334}
Kyle Hsu, Sergey Levine, and Chelsea Finn.
\newblock Unsupervised learning via meta-learning.
\newblock {\em CoRR}, abs/1810.02334, 2018.

\bibitem{hsu2017learning}
Yen-Chang Hsu, Zhaoyang Lv, and Zsolt Kira.
\newblock Learning to cluster in order to transfer across domains and tasks.
\newblock {\em arXiv preprint arXiv:1711.10125}, 2017.

\bibitem{huang2017cross}
Xin Huang, Yuxin Peng, and Mingkuan Yuan.
\newblock Cross-modal common representation learning by hybrid transfer
  network.
\newblock In {\em Proceedings of the 26th International Joint Conference on
  Artificial Intelligence}, pages 1893--1900, 2017.

\bibitem{DBLP:journals/corr/HuhAE16}
Mi{-}Young Huh, Pulkit Agrawal, and Alexei~A. Efros.
\newblock What makes imagenet good for transfer learning?
\newblock {\em CoRR}, abs/1608.08614, 2016.

\bibitem{DBLP:journals/corr/abs-0907-1815}
Hal~Daum{\'{e}} III.
\newblock Frustratingly easy domain adaptation.
\newblock {\em CoRR}, abs/0907.1815, 2009.

\bibitem{jawanpuria2020geometry}
Pratik Jawanpuria, Mayank Meghwanshi, and Bamdev Mishra.
\newblock Geometry-aware domain adaptation for unsupervised alignment of word
  embeddings.
\newblock {\em arXiv preprint arXiv:2004.08243}, 2020.

\bibitem{k-m-etal-2018-learning}
Annervaz K~M, Somnath Basu Roy~Chowdhury, and Ambedkar Dukkipati.
\newblock Learning beyond datasets: Knowledge graph augmented neural networks
  for natural language processing.
\newblock In {\em Proceedings of the 2018 Conference of the North {A}merican
  Chapter of the Association for Computational Linguistics: Human Language
  Technologies, Volume 1 (Long Papers)}, 2018.

\bibitem{DBLP:journals/corr/abs-1902-10644}
Mikhail Khodak, Maria{-}Florina Balcan, and Ameet Talwalkar.
\newblock Provable guarantees for gradient-based meta-learning.
\newblock {\em CoRR}, abs/1902.10644, 2019.

\bibitem{DBLP:journals/corr/KrishnaZGJHKCKL16}
Ranjay Krishna, Yuke Zhu, Oliver Groth, Justin Johnson, Kenji Hata, Joshua
  Kravitz, Stephanie Chen, Yannis Kalantidis, Li{-}Jia Li, David~A. Shamma,
  Michael~S. Bernstein, and Fei{-}Fei Li.
\newblock Visual genome: Connecting language and vision using crowdsourced
  dense image annotations.
\newblock {\em CoRR}, abs/1602.07332, 2016.

\bibitem{krizhevsky2009learning}
Alex Krizhevsky, Geoffrey Hinton, et~al.
\newblock Learning multiple layers of features from tiny images.
\newblock 2009.

\bibitem{kumar2018knowledge}
Anurag {Kumar}, Maksim {Khadkevich}, and Christian {Fugen}.
\newblock Knowledge transfer from weakly labeled audio using convolutional
  neural network for sound events and scenes.
\newblock In {\em 2018 IEEE International Conference on Acoustics, Speech and
  Signal Processing (ICASSP)}, pages 326--330, 2018.

\bibitem{larsen2016autoencoding}
Anders Boesen~Lindbo Larsen, S{\o}ren~Kaae S{\o}nderby, Hugo Larochelle, and
  Ole Winther.
\newblock Autoencoding beyond pixels using a learned similarity metric.
\newblock In {\em International conference on machine learning}, pages
  1558--1566. PMLR, 2016.

\bibitem{li2019santlr}
Xinjian Li, Zhong Zhou, Siddharth Dalmia, Alan~W Black, and Florian Metze.
\newblock Santlr: Speech annotation toolkit for low resource languages.
\newblock {\em arXiv preprint arXiv:1908.01067}, 2019.

\bibitem{liang2020think}
Paul~Pu Liang, Terrance Liu, Liu Ziyin, Ruslan Salakhutdinov, and
  Louis-Philippe Morency.
\newblock Think locally, act globally: Federated learning with local and global
  representations.
\newblock {\em arXiv preprint arXiv:2001.01523}, 2020.

\bibitem{liang2019learning}
Paul~Pu Liang, Zhun Liu, Yao-Hung~Hubert Tsai, Qibin Zhao, Ruslan
  Salakhutdinov, and Louis-Philippe Morency.
\newblock Learning representations from imperfect time series data via tensor
  rank regularization.
\newblock In {\em Proceedings of the 57th Annual Meeting of the Association for
  Computational Linguistics}, pages 1569--1576, 2019.

\bibitem{liang2018multimodal}
Paul~Pu Liang, Amir Zadeh, and Louis-Philippe Morency.
\newblock Multimodal local-global ranking fusion for emotion recognition.
\newblock In {\em Proceedings of the 20th ACM International Conference on
  Multimodal Interaction}, pages 472--476, 2018.

\bibitem{Miller:1995:WLD:219717.219748}
George~A. Miller.
\newblock Wordnet: A lexical database for english.
\newblock {\em Commun. ACM}, 38(11):39--41, Nov. 1995.

\bibitem{yummly_dataset}
W. {Min}, S. {Jiang}, J. {Sang}, H. {Wang}, X. {Liu}, and L. {Herranz}.
\newblock Being a supercook: Joint food attributes and multimodal content
  modeling for recipe retrieval and exploration.
\newblock {\em IEEE Transactions on Multimedia}, 19(5):1100--1113, 2017.

\bibitem{munkhdalai2017meta}
Tsendsuren Munkhdalai and Hong Yu.
\newblock Meta networks.
\newblock In {\em Proceedings of the 34th International Conference on Machine
  Learning-Volume 70}, pages 2554--2563. JMLR. org, 2017.

\bibitem{nichol2018reptile}
Alex Nichol and John Schulman.
\newblock Reptile: a scalable metalearning algorithm.
\newblock {\em arXiv preprint arXiv:1803.02999}, 2:2, 2018.

\bibitem{DBLP:journals/corr/abs-1808-00177}
OpenAI, Marcin Andrychowicz, Bowen Baker, Maciek Chociej, Rafal
  J{\'{o}}zefowicz, Bob McGrew, Jakub~W. Pachocki, Jakub Pachocki, Arthur
  Petron, Matthias Plappert, Glenn Powell, Alex Ray, Jonas Schneider, Szymon
  Sidor, Josh Tobin, Peter Welinder, Lilian Weng, and Wojciech Zaremba.
\newblock Learning dexterous in-hand manipulation.
\newblock {\em CoRR}, abs/1808.00177, 2018.

\bibitem{DBLP:journals/corr/abs-1806-03560}
Akila Pemasiri, Kien Nguyen, Sridha Sridharan, and Clinton Fookes.
\newblock Semantic correspondence: {A} hierarchical approach.
\newblock {\em CoRR}, abs/1806.03560, 2018.

\bibitem{perez-rosas_utterance-level_2013}
Veronica Perez-Rosas, Rada Mihalcea, and Louis-Philippe Morency.
\newblock Utterance-{Level} {Multimodal} {Sentiment} {Analysis}.
\newblock In {\em Association for {Computational} {Linguistics} ({ACL})},
  Sofia, Bulgaria, Aug. 2013.

\bibitem{piczak2015esc}
Karol~J Piczak.
\newblock Esc: Dataset for environmental sound classification.
\newblock In {\em Proceedings of the 23rd ACM international conference on
  Multimedia}, pages 1015--1018, 2015.

\bibitem{DBLP:journals/corr/RajNT15a}
Anant Raj, Vinay~P. Namboodiri, and Tinne Tuytelaars.
\newblock Subspace alignment based domain adaptation for {RCNN} detector.
\newblock {\em CoRR}, abs/1507.05578, 2015.

\bibitem{ravi2016optimization}
Sachin Ravi and Hugo Larochelle.
\newblock Optimization as a model for few-shot learning.
\newblock 2016.

\bibitem{ridnik2020tresnet}
Tal Ridnik, Hussam Lawen, Asaf Noy, and Itamar Friedman.
\newblock Tresnet: High performance gpu-dedicated architecture.
\newblock {\em arXiv preprint arXiv:2003.13630}, 2020.

\bibitem{sahoo2019metalearning}
Doyen Sahoo, Hung Le, Chenghao Liu, and Steven C.~H. Hoi.
\newblock Meta-learning with domain adaptation for few-shot learning under
  domain shift, 2019.

\bibitem{DBLP:journals/corr/abs-1902-09492}
Tal Schuster, Ori Ram, Regina Barzilay, and Amir Globerson.
\newblock Cross-lingual alignment of contextual word embeddings, with
  applications to zero-shot dependency parsing.
\newblock {\em CoRR}, abs/1902.09492, 2019.

\bibitem{snell2017prototypical}
Jake Snell, Kevin Swersky, and Richard Zemel.
\newblock Prototypical networks for few-shot learning.
\newblock In {\em Advances in neural information processing systems}, pages
  4077--4087, 2017.

\bibitem{10.5555/2999611.2999716}
Richard Socher, Milind Ganjoo, Christopher~D. Manning, and Andrew~Y. Ng.
\newblock Zero-shot learning through cross-modal transfer.
\newblock In {\em Proceedings of the 26th International Conference on Neural
  Information Processing Systems - Volume 1}, NIPS’13, 2013.

\bibitem{DBLP:journals/corr/abs-1808-01974}
Chuanqi Tan, Fuchun Sun, Tao Kong, Wenchang Zhang, Chao Yang, and Chunfang Liu.
\newblock A survey on deep transfer learning.
\newblock {\em CoRR}, abs/1808.01974, 2018.

\bibitem{DBLP:journals/corr/abs-1710-08347}
Yao{-}Hung~Hubert Tsai and Ruslan Salakhutdinov.
\newblock Improving one-shot learning through fusing side information.
\newblock {\em CoRR}, abs/1710.08347, 2017.

\bibitem{tzeng2017adversarial}
Eric Tzeng, Judy Hoffman, Kate Saenko, and Trevor Darrell.
\newblock Adversarial discriminative domain adaptation.
\newblock In {\em Proceedings of the IEEE Conference on Computer Vision and
  Pattern Recognition}, pages 7167--7176, 2017.

\bibitem{DBLP:journals/corr/TzengHZSD14}
Eric Tzeng, Judy Hoffman, Ning Zhang, Kate Saenko, and Trevor Darrell.
\newblock Deep domain confusion: Maximizing for domain invariance.
\newblock {\em CoRR}, abs/1412.3474, 2014.

\bibitem{vinyals2016matching}
Oriol Vinyals, Charles Blundell, Timothy Lillicrap, Daan Wierstra, et~al.
\newblock Matching networks for one shot learning.
\newblock In {\em Advances in neural information processing systems}, pages
  3630--3638, 2016.

\bibitem{10.5555/2283516.2283652}
Chang Wang and Sridhar Mahadevan.
\newblock Heterogeneous domain adaptation using manifold alignment.
\newblock In {\em Proceedings of the Twenty-Second International Joint
  Conference on Artificial Intelligence - Volume Volume Two}, IJCAI’11, page
  1541–1546. AAAI Press, 2011.

\bibitem{wang2019cross}
Zirui Wang, Jiateng Xie, Ruochen Xu, Yiming Yang, Graham Neubig, and Jaime
  Carbonell.
\newblock Cross-lingual alignment vs joint training: A comparative study and a
  simple unified framework.
\newblock {\em arXiv preprint arXiv:1910.04708}, 2019.

\bibitem{NIPS2019_8731}
Chen Xing, Negar Rostamzadeh, Boris Oreshkin, and Pedro~O O.~Pinheiro.
\newblock Adaptive cross-modal few-shot learning.
\newblock In H. Wallach, H. Larochelle, A. Beygelzimer, F. d\textquotesingle
  Alch\'{e}-Buc, E. Fox, and R. Garnett, editors, {\em Advances in Neural
  Information Processing Systems 32}. 2019.

\bibitem{10.1007/978-3-319-68783-4_35}
Zhenguo Yang, Min Cheng, Qing Li, Yukun Li, Zehang Lin, and Wenyin Liu.
\newblock Cross-domain and cross-modality transfer learning for multi-domain
  and multi-modality event detection.
\newblock In Athman Bouguettaya, Yunjun Gao, Andrey Klimenko, Lu Chen,
  Xiangliang Zhang, Fedor Dzerzhinskiy, Weijia Jia, Stanislav~V. Klimenko, and
  Qing Li, editors, {\em Web Information Systems Engineering -- WISE 2017},
  pages 516--523, Cham, 2017. Springer International Publishing.

\bibitem{zadeh2020foundations}
Amir Zadeh, Paul~Pu Liang, and Louis-Philippe Morency.
\newblock Foundations of multimodal co-learning.
\newblock {\em Information Fusion}, 64:188--193, 2020.

\bibitem{zadeh2018multimodal}
AmirAli~Bagher Zadeh, Paul~Pu Liang, Soujanya Poria, Erik Cambria, and
  Louis-Philippe Morency.
\newblock Multimodal language analysis in the wild: Cmu-mosei dataset and
  interpretable dynamic fusion graph.
\newblock In {\em ACL}, 2018.

\bibitem{zhang2019category}
Qiming Zhang, Jing Zhang, Wei Liu, and Dacheng Tao.
\newblock Category anchor-guided unsupervised domain adaptation for semantic
  segmentation.
\newblock In {\em Advances in Neural Information Processing Systems}, pages
  433--443, 2019.

\end{thebibliography}
\clearpage

\appendix

\addtocounter{definition}{-2}
\addtocounter{proposition}{-1}

\section*{Appendix}

\vspace{-1mm}
\section{Formalizing Cross-modal Generalization}
\label{formalize_supp}
\vspace{-1mm}

In the main text, we have established, with some degree of formality, a framework for studying cross-modal generalization. In this section, we extend and add in details surrounding the discussion in the main text. This will allow us to understanding our method in a more precise manner.

\textit{Cross-modal generalization} is a learning paradigm to quickly perform new tasks in a target modality despite being trained on a different source modality. To formalize this paradigm, we build on the definition of meta-learning~\cite{hospedales2020meta} and generalize it to study multiple input modalities.
The goal of meta-learning can be broadly defined as using labeled data for existing source tasks to learn representations that enable fast learning on unseen target tasks~\cite{DBLP:journals/corr/abs-1902-10644}.
To reason over multiple modalities and tasks, we start by defining $M$ different heterogeneous input spaces (modalities) and $N$ different label spaces (tasks).
We denote a modality by an index $m\in \{1,...,M\}$ and a task by $n\in \{1,...,N\}$.

Each classification problem $\mathcal{T}(m,n)$ is defined as a triplet with a modality, task, plus a joint distribution: $\mathcal{T}(m,n) = (\mathcal{X}_m, \mathcal{Y}_n, p_{m,n}(x, y))$.
$\mathcal{X}_{m}$ denotes the input space and $\mathcal{Y}_{n}$ the label space sampled from a distribution $p(m,n) := p(\mathcal{X}_{m}, \mathcal{Y}_{n})$ given by a marginal over the entire \textit{meta-distribution}, $p(x_1,..., x_M, y_1, ...y_N, \mathcal{X}_{m_1},...\mathcal{X}_{m_M}, \mathcal{Y}_{n_1},...\mathcal{Y}_{n_N})$. The meta-distribution gives the underlying relationships between all modalities and tasks through a hierarchical generative process $m_i \sim p(m), n_j \sim p(n)$: first picking a modality and task $(m_i,n_j)$ from priors $p(m)$ and $p(n)$ over input and output spaces, before drawing data $x_i$ from $\mathcal{X}_{m_i}$ and labels $y_j$ from $\mathcal{Y}_{n_j}$.
Within each classification problem is also an underlying pairing function mapping inputs to labels through $p_{m,n}(x,y) := p(x,y | m,n)$ for all $x\in \mathcal{X}_{m},\ y\in \mathcal{Y}_{n}$ representing the true data labeling process. Note that in practice $p_{m,n}(x,y)$ is never known but instead represented as (modality, label) pairs as collected and annotated as real-world datasets.

To account for generalization over modalities and tasks, cross-modal generalization involves learning a single function $f_w$ with parameters $w$ over the meta-distribution with the following objective:
\begin{definition}
    \underline{Cross-modal generalization} is a maximization problem given by
    \begin{equation}
        \label{eq: meta-learning generalization_supp}
        \argmax_w \mathcal{L}[f_w] := \argmax_w \underset{\substack{m,n\sim p(m,n) \\ x,y\sim p_{m,n}(x,y)}}{\mathbb{E}} \log\left[\frac{f_w(x,y,m,n)}{p(x,y| m,n)}\right].
    \end{equation}
    When $p(n)$ is a delta distribution, we say that the problem is \underline{single task}; otherwise, it is \underline{multi-task}. $p(m)$ is any arbitrary distribution over the source domains.
\end{definition}
We call eq~\eqref{eq: meta-learning generalization_supp} the generalization loss and the goal of any model we consider is to minimize this loss. Notice that this loss is lower bounded by $0$, and is achievable when $f_w(x,y,m,n) = p(x,y| m,n)$. A model $f_w$ that achieves $0$ loss in eq~\eqref{eq: meta-learning generalization_supp} is said to achieve \underline{perfect generalization}.

\vspace{-0mm}
\subsection{Cross-modal Few-shot learning}
\vspace{-0mm}

In practice, the space between modalities and tasks is only \textit{partially observed}: $p(x,y|m,n)$ is only observed for certain modalities and tasks (e.g. labeled classification tasks for images~\cite{deng2009imagenet}, or paired data across image, text, and audio in online videos~\cite{45619}). For other modality-task pairs, we can only obtain inaccurate estimates $q(x,y|m,n)$, often due to having only \textit{limited labeled data}.
We are now ready to define a \textit{few-shot learning} problem.
\begin{definition}
     Let $\mathcal{M}$ be a subset of all the possible pairings of modality and task spaces. A meta-learning problem is said to be (partially) \underline{low resource} if 
     for all $m,\ n\in \mathcal{M}$, $p(x,y|m,n)$ is not known exactly, and has to be estimated using $q(x,y|m,n)\neq p(x,y|m,n)$.
\end{definition}
Therefore, the subset $\mathcal{M}$ can be called the low-resource subset, and any task associated with $\mathcal{M}$ is a low-resource task. Note that this definition is equivalent to a situation where we have infinitely many data points for the high resource tasks, and a finite number of data points for the low-resource tasks. 
In practice, $q(x,y|m,n)$ is an (imperfect) estimation of $p(x,y|m,n)$ due to limited labeled data. Mathematically, for a few-shot meta-learning problem, the optimization objective becomes
{\fontsize{9.5}{12}\selectfont
\begin{align}
    \label{eq: supervised meta-learning}
    &\argmax_w \mathcal{L}_q[f_w] := \nonumber \\
    &\argmax_w \underset{\substack{m,n\\ \sim p(m,n)}}{\mathbb{E}}  \underbrace{\bigg\{ \underset{\substack(m,n)\notin \mathcal{M}}{\mathbf{1}} \underset{\substack{x,y\\ \sim p_{m,n}(x,y)}}{\mathbb{E}} \log\left[\frac{f_w(x,y,m,n)}{p(x,y| m,n)}\right]}_{\text{high resource subset}} \nonumber \\
    &+ \underbrace{\underset{\substack(m,n) \in \mathcal{M}}{\mathbf{1}} \underset{\substack{x,y\\ \sim q_{m,n}(x,y)}}{\mathbb{E}} \log\left[\frac{f_w(x,y,m,n)}{q(x,y| m,n)}\right]}_{\text{low resource subset} }\bigg\} ,
\end{align}
}where $\mathbf{1}$ is the indicator function. This new optimization objective no longer matches the generalization objective $\mathcal{L}$ in eq~\eqref{eq: meta-learning generalization_supp}. The minimizer of this equation is $f_w(x,y,m,n) = q(x,y| m,n)$, which has an generalization error $\mathrm{Pr}\{(m,n)\in \mathcal{M}\}\textrm{KL}(p;q)$, where $\textrm{KL}(\cdot;\cdot)$ is the KL-divergence measuring how inaccurate the real-life estimates $q$ are due to limited labeled data.

What is the minimal extra supervision required to perform cross-modal generalization under only partial observability? 
To answer this question, we first define the minimum requirements on observed data, which we call the \textit{minimum visibility assumption}:
\begin{assumption}
    (Minimum visibility) For every task $n$, there is at least one domain $m$ such that $p(x,y|m,n)$ is known. Likewise, for every domain $m$, there is at least one task $n$ such $p(x,y|m,n)$ is known. All the single variable marginal distributions $p(x)$, $p(y)$ are also known.
\end{assumption}

\begin{figure}[tbp]
\centering
\vspace{-2mm}
\includegraphics[width=0.7\linewidth]{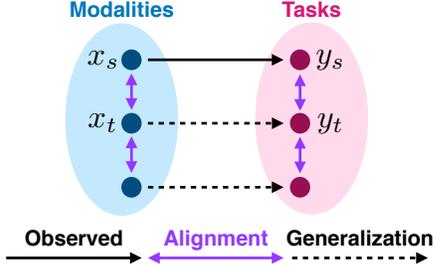}
\caption{A modality-task graph contains a subset of observed edges through labeled datasets for specific modalities and tasks. Generalizing to the remaining modalities and tasks (dotted edge) requires bridging modalities through alignment.\vspace{-2mm}}
\label{graph_supp}
\end{figure}

In practice, we say that a distribution is known if it can be accurately estimated.
This is the minimum assumption required to ensure that all modalities and tasks are accessible. 
It is helpful to think about this \textit{partial observability} as a bipartite graph $G=(V_x,V_y,E)$ between a modality set $V_x$ and task set $V_y$ (see Figure~\ref{graph_supp}). A solid directed edge from $u \in V_x$ to $v \in V_y$ represents learning a classifier from modality $u$ for task $v$ given an abundance of observed labeled data, which incurs negligible generalization error.
Since it is unlikely for all edges between $V_x$ and $V_y$ to exist, define the \textit{low-resource subset} $\mathcal{M}$ as the complement of $E$ in $V_x \times V_y$. $\mathcal{M}$ represents the set of low-resource modalities and tasks where it is difficult to obtain labeled data.
The focus of cross-modal generalization is to learn a classifier in $\mathcal{M}$ as denoted by a dashed edge. In contrast to solid edges, the lack of data in $\mathcal{M}$ incurs large error along dashed edges. It is helpful to differentiate solid vs dashed edges by writing them as weighted edges $(u, v, \epsilon)$, where $\epsilon$ denotes error incurred.
The visibility assumption says that there is at least a solid in/out edge for every vertex in $V_m$ and $V_n$.



\vspace{-0mm}
\subsection{Cross-modal Alignment}
\label{theory:alignment}
\vspace{-0mm}

Therefore, the challenge in cross-modal generalization amounts to finding the path of lowest cumulative error between an input target modality $x_t \in V_x$ and output task $y_t \in V_y$ in $\mathcal{M}$. The key insight is to leverage \textit{cross-modal information} to ``bridge'' modalities that are each labeled for only a subset of tasks (see purple edges in Figure~\ref{graph_supp}). We model cross-modal information as $p(x_s, x_t)$, i.e. \textit{alignment} between modalities $x_s$ and $x_t$, where $x_s$ is a source modality with high-resource data and labels $(x_s, y_s)$. When there is an abundance of paired data $(x_s, x_t)$ (solid purple edge), we say that \textit{strong} alignment exists; otherwise, only \textit{weak} alignment exits. Since strong alignment incurs negligible error in estimating $p(x_s, x_t)$, the alternative \textit{cross-modal path} $P = \{(x_t, x_s), (x_s, y_s), (y_s, y_t)\}$ might link $x_t$ and $y_t$ with \textit{lower} weighted error and is preferable to direct low-resource training for the dashed edge $(x_t, y_t)$. When only weak alignment is available, a trade-off emerges and one has to choose between the error induced by direct low-resource training and the error induced by weak alignment. $(y_s, y_t)$ models relationships across source and target tasks using approaches such as multi-task~\cite{caruana1997multitask} or meta-learning~\cite{finn2017model}. More formally,
\begin{definition}
    Let $p(x_i, x_j)$ be known for $x_i\in \mathcal{D}_{m_i}^{x}$, $x_j\in \mathcal{D}_{m_j}^{x}$ and $i\neq j$. If both $p(x_i|x_j)$ and $p(x_j|x_i)$ are delta distributions, i.e., if there is a one-to-one mapping between $x_i$ and $x_j$, we say that there is a \underline{strong alignment} between modality $m_i$ and $m_j$. Otherwise, there is only \underline{weak alignment}.
\end{definition}
We now show that strong alignment across modalities can achieve optimal generalization error for tasks in the low-resource subset $\mathcal{M}$.
\begin{proposition}
    \label{prop: strong align_supp}
    (Benefit of strong alignment). Let all the modalities be pairwise strongly-aligned, then we can define a surrogate loss function $\Tilde{\mathcal{L}}[f_w]$ such that $\mathcal{L}[\argmin_{f_w} \Tilde{\mathcal{L}}[f_w] ] = 0$.
\end{proposition}

\textit{Proof.} Let $\mathcal{T}_{s,t}$ be in the low-resource set, where we only know $q(x_{t},y)$. We want to show that we can recover $p(x_t,y)$ from alignment information. By the assumption of visibility, for task $t$, there is a strongly aligned modality $s \neq t$ for which we know $p(x_{s}, x_{t})$. By Bayes' rule $p(x_{t}, x_{s}, y) = p(x_{t}| x_{s}, y) p(x_{s}, y)$, but $x_{s}$ is conditionally independent of $y$ if $x_{t}$ is known due to the existence of one-to-one mapping between them. Therefore, we can calculate $p(x_{t}, x_{s}, y) = p(x_{t}| x_{s}) p(x_{s}, y)$ recover the desired label $p(x_{t}, y) = \int p(x_{s}, x_{t}, y) \ d x_{s}$. Now we can replace $q(x_t,y)$ by the recovered $p(x_t,y)$ in the loss function, thus achieving perfect generalization on this task. $\square$

This implies that if strong alignment is achievable, then one can achieve perfect generalization in the low-resource subset $\mathcal{M}$.
We also note that a key property we used in the proof is that $p(x_t|x_s) = p(x_t|x_s,y)$. For weak alignment, this property does not hold and perfect generalization is no longer achievable, and one needs to tradeoff the error induced by weak alignment with the error from minimizing $q$ directly (i.e. few-shot supervised learning).
We further explain and qualitatively analyze this trade-off in Appendix~\ref{tradeoff_supp}.

Therefore, unlabeled cross-modal information $p(x_s, x_t)$ allows us to bridge modalities that are each labeled for only a subset of tasks and achieve cross-modal generalization to new modalities and tasks in $\mathcal{M}$. In practice, however, $p(x_s, x_t)$ is unknown and needs to be estimated from data, and is the basis for our proposed \names\ approach to estimate $p(x_s, x_t)$ from data and meta-learning to model $(y_s, y_t)$.

\vspace{-1mm}
\subsection{Concerning Weak Alignment}
\vspace{-1mm}

For weak alignment, this property may not hold and perfect generalization may not be achievable. Therefore, one needs to tradeoff the error induced by weak alignment with the error from minimizing $q$ directly (i.e. few-shot supervised learning).
This does not necessarily mean that weak alignment will hurt generalization: if $p(x_t|x_s,y) = p(x_t|x_s)$ holds, then perfect generalization can still be achieved.
Of course, one might differentiate between the \textit{perfect weak alignment} problem, where the statement $p(x_t|x_s,y) = p(x_t|x_s)$ holds (or, requiring one additional assumption) and \textit{proper weak alignment}, where it does not. One can therefore prove the following corollary.
\begin{corollary}
    Assuming perfect weak alignment, one can achieve perfect generalization error.
\end{corollary}
The proof follows directly from Proposition~\ref{prop: strong align}.

\vspace{-1mm}
\section{Experimental Details}
\label{details_supp}
\vspace{-1mm}

The code for running our experiments can be found in the supplementary material. We also provide some experimental details below. Since there are no established benchmarks in cross-modal generalization, we create our own by merging and preprocessing several multimodal datasets. We believe that these two benchmarks for assessing cross-modal generalization (image to audio and text to speech) will also be useful to the broader research community and hence we also open-source all data and data processing code.

\vspace{-1mm}
\subsection{Text to Image}
\label{text_image_supp}
\vspace{-1mm}

\textbf{Data:} We use the Yummly-28K dataset~\cite{yummly_dataset} which contains parallel text descriptions and images of recipes. We create classification labels from the metadata by concatenating the meal type and cuisine, yielding $44$ distinct classes. The large number of recipes and shared concepts between text and image makes it an ideal testbed for cross-modal generalization.
We used a ResNet pretrained on ImageNet~\cite{deng2009imagenet} to encode the images, pretrained BERT encoder~\cite{devlin2018bert} for text, and a shared network for prediction.

\textbf{Hyperparameters:} We show the hyperparameters used in Table~\ref{recipe_params}.

\begin{table}[t]
\fontsize{9}{11}\selectfont
\centering
\caption{Table of hyperparameters for generalization experiments on text to image task. Batchsize $4/8/16$ indicates the batchsize used for $1/5/10$-shot experiments respectively.\vspace{2mm}}
\setlength\tabcolsep{3.5pt}
\begin{tabular}{l | c | c}
\Xhline{3\arrayrulewidth}
Model & Parameter & Value \\
\Xhline{0.5\arrayrulewidth}
\multirow{14}{*}{Text Encoder}
& Shared layer size & $256$ \\
& Batchsize & $4/8/16$ \\
& Activation & ReLU \\
& Meta Optimizer & SGD \\
& Optimizer & Adam \\
& Meta Learning rate & $1e-1$ \\
& Align Learning rate & $1e-3$ \\
& Classifier Learning rate & $1e-3$\\
& Iterations & $800$\\
& Number of evaluation tasks & 16\\
& Loss Margin & $0.1$\\
& Width Factor & $1.3$ \\
& Number of Layers & $4$ \\
& Blocks Per Layer & $4, 5, 24, 3$ \\
\Xhline{3\arrayrulewidth}
\end{tabular}

\vspace{2mm}

\begin{tabular}{l | c | c}
\Xhline{3\arrayrulewidth}
Model & Parameter & Value \\
\Xhline{0.5\arrayrulewidth}
\multirow{14}{*}{Image Encoder}
& Shared layer size & $256$ \\
& Batchsize & $4/8/16$ \\
& Activation & ReLU \\
& Meta Optimizer & SGD \\
& Optimizer & Adam \\
& Meta Learning rate & $1e-1$ \\
& Align Learning rate & $1e-3$ \\
& Classifier Learning rate & $1e-3$\\
& Iterations & $800$\\
& Number of evaluation tasks & 16\\
& Loss Margin & $0.1$\\
& Intermediate Pooling Function & Max \\
& Final Pooling Function & Average \\
& Stride & 1 \\
\Xhline{3\arrayrulewidth}
\end{tabular}

\vspace{2mm}

\begin{tabular}{l | c | c}
\Xhline{3\arrayrulewidth}
Model & Parameter & Value \\
\Xhline{0.5\arrayrulewidth}
\multirow{13}{*}{Image Classifier}
& Num hidden layers & $1$ \\
& Hidden layer size & $256$ \\
& Batchsize & $4/8/16$ \\
& Activation & ReLU \\
& Meta Optimizer & SGD \\
& Optimizer & Adam \\
& Meta Learning rate & $1e-1$ \\
& Align Learning rate & $1e-3$ \\
& Classifier Learning rate & $1e-3$\\
& Iterations & $800$\\
& Number of evaluation tasks & 16\\
& Loss & MSE\\
& Teacher forcing rate & $0.5$ \\
\Xhline{3\arrayrulewidth}
\end{tabular}
\vspace{-4mm}
\label{recipe_params}
\end{table}

\vspace{-1mm}
\subsection{Image to Audio}
\label{image_audio_supp}
\vspace{-1mm}

\begin{figure*}[tbp]
\centering
\vspace{-4mm}
\includegraphics[width=\linewidth]{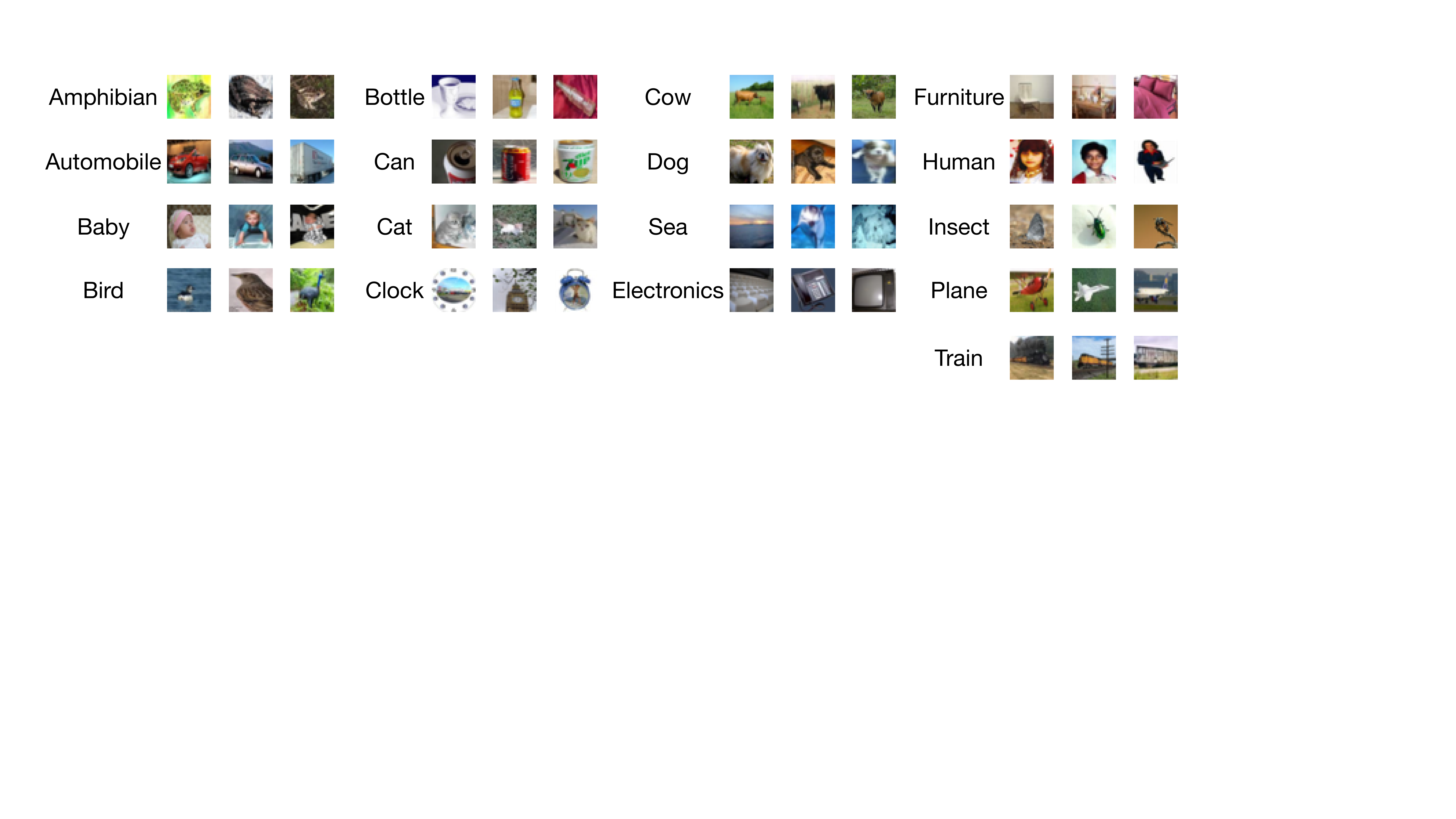}
\caption{The $17$ concepts shared across image and audio classification tasks that were used for weak alignment. Note that we only show the images - the audio spectrograms make up the second modality in each weak cluster.\vspace{-4mm}}
\label{concepts}
\end{figure*}

\textbf{Data:} To construct our generalization dataset, we combine $100$ classes from CIFAR-$100$ and $10$ classes from CIFAR-$10$~\cite{krizhevsky2009learning} to form $110$ image classes, as well as $50$ audio classes from ESC-$50$~\cite{piczak2015esc}. The tasks across these modalities are different (i.e. different classification problems) which requires cross-modal generalization. To bridge these two modalities with partially related label spaces, we define $17$ shared classes across the $2$ datasets for weak concept alignment. We show the $17$ clustered concepts we used for weak alignment in Figure~\ref{concepts}. These clusters are obtained by mapping similar classes between the datasets using similarities from WordNet~\cite{Miller:1995:WLD:219717.219748} and text cooccurrence.
The number of shared classes in train, val, and test, respectively is $12$, $8$, and $9$, and the number of samples is $920$, $580$, $580$, respectively.

\textbf{Hyperparameters:} We show the hyperparameters used in Table~\ref{image_audio_params}.

\begin{table}[t]
\fontsize{9}{11}\selectfont
\centering
\caption{Table of hyperparameters for generalization experiments on image to audio task. Batchsize $4/8/16$ indicates the batchsize used for $1/5/10$-shot experiments respectively.\vspace{2mm}}
\setlength\tabcolsep{3.5pt}
\begin{tabular}{l | c | c}
\Xhline{3\arrayrulewidth}
Model & Parameter & Value \\
\Xhline{0.5\arrayrulewidth}
\multirow{14}{*}{Image Encoder}
& Shared layer size & $256$ \\
& Batchsize & $4/8/16$ \\
& Activation & ReLU \\
& Meta Optimizer & SGD \\
& Optimizer & Adam \\
& Meta Learning rate & $1e-1$ \\
& Align Learning rate & $1e-3$ \\
& Classifier Learning rate & $1e-3$\\
& Iterations & $800$\\
& Number of evaluation tasks & 16\\
& Loss Margin & $0.1$\\
& Width Factor & $1.3$ \\
& Number of Layers & $4$ \\
& Blocks Per Layer & $4, 5, 24, 3$ \\
\Xhline{3\arrayrulewidth}
\end{tabular}

\vspace{2mm}

\begin{tabular}{l | c | c}
\Xhline{3\arrayrulewidth}
Model & Parameter & Value \\
\Xhline{0.5\arrayrulewidth}
\multirow{14}{*}{Audio Encoder}
& Shared layer size & $256$ \\
& Batchsize & $4/8/16$ \\
& Activation & ReLU \\
& Meta Optimizer & SGD \\
& Optimizer & Adam \\
& Meta Learning rate & $1e-1$ \\
& Align Learning rate & $1e-3$ \\
& Classifier Learning rate & $1e-3$\\
& Iterations & $800$\\
& Number of evaluation tasks & 16\\
& Loss Margin & $0.1$\\
& Intermediate Pooling Function & Max \\
& Final Pooling Function & Average \\
& Stride & 1 \\
\Xhline{3\arrayrulewidth}
\end{tabular}

\vspace{2mm}

\begin{tabular}{l | c | c}
\Xhline{3\arrayrulewidth}
Model & Parameter & Value \\
\Xhline{0.5\arrayrulewidth}
\multirow{13}{*}{Audio Classifier}
& Num hidden layers & $1$ \\
& Hidden layer size & $256$ \\
& Batchsize & $4/8/16$ \\
& Activation & ReLU \\
& Meta Optimizer & SGD \\
& Optimizer & Adam \\
& Meta Learning rate & $1e-1$ \\
& Align Learning rate & $1e-3$ \\
& Classifier Learning rate & $1e-3$\\
& Iterations & $800$\\
& Number of evaluation tasks & 16\\
& Loss & MSE\\
& Teacher forcing rate & $0.5$ \\
\Xhline{3\arrayrulewidth}
\end{tabular}
\vspace{-4mm}
\label{image_audio_params}
\end{table}

\begin{table}[t]
\fontsize{9}{11}\selectfont
\centering
\caption{Table of hyperparameters for generalization experiments on text to speech task. Batchsize $4/8/16$ indicates the batchsize used for $1/5/10$-shot experiments respectively.\vspace{2mm}}
\setlength\tabcolsep{3.5pt}
\begin{tabular}{l | c | c}
\Xhline{3\arrayrulewidth}
Model & Parameter & Value \\
\Xhline{0.5\arrayrulewidth}
\multirow{14}{*}{Text Encoder}
& Bidirectional & True \\
& Embedding dim & $256$ \\
& Num hidden layers & $1$ \\
& Hidden layer size & $256$ \\
& Batchsize & $4/8/16$ \\
& Activation & ReLU \\
& Meta Optimizer & SGD \\
& Optimizer & Adam \\
& Meta Learning rate & $1e-1$ \\
& Align Learning rate & $1e-3$ \\
& Classifier Learning rate & $1e-4$\\
& Iterations & $800$\\
& Loss Margin & $0.1$\\
& Number of evaluation tasks & $16$\\
\Xhline{3\arrayrulewidth}
\end{tabular}

\vspace{2mm}

\begin{tabular}{l | c | c}
\Xhline{3\arrayrulewidth}
Model & Parameter & Value \\
\Xhline{0.5\arrayrulewidth}
\multirow{13}{*}{Speech Encoder}
& Embedding dim & $40$ \\
& Num hidden layers & $2$ \\
& Hidden layer size & $256$ \\
& Batchsize & $4/8/16$ \\
& Activation & ReLU \\
& Meta Optimizer & SGD \\
& Optimizer & Adam \\
& Meta Learning rate & $1e-1$ \\
& Align Learning rate & $1e-3$ \\
& Classifier Learning rate & $1e-4$\\
& Iterations & $800$\\
& Loss Margin & $0.1$\\
& Number of evaluation tasks & $16$\\
\Xhline{3\arrayrulewidth}
\end{tabular}

\vspace{2mm}

\begin{tabular}{l | c | c}
\Xhline{3\arrayrulewidth}
Model & Parameter & Value \\
\Xhline{0.5\arrayrulewidth}
\multirow{13}{*}{Speech Classifier}
& Num hidden layers & $1$ \\
& Hidden layer size & $256$ \\
& Batchsize & $4/8/16$ \\
& Activation & ReLU \\
& Meta Optimizer & SGD \\
& Optimizer & Adam \\
& Meta Learning rate & $1e-1$ \\
& Align Learning rate & $1e-3$ \\
& Classifier Learning rate & $1e-4$\\
& Iterations & $800$\\
& Loss & MSE\\
& Teacher forcing rate & $0.5$ \\
& Number of evaluation tasks & $16$\\
\Xhline{3\arrayrulewidth}
\end{tabular}
\vspace{-0mm}
\label{tts_params}
\end{table}

\vspace{-1mm}
\subsection{Text to Speech}
\label{tts_supp}
\vspace{-1mm}

\textbf{Data:} The dataset is composed of paired text-speech data from a $99$-language subset of the Wilderness dataset~\cite{8683536}. The dataset was collected using text and speech from the Bible. We preprocessed the data so that every language corresponded to a different set of chapters, maximizing the independence between datapoints across languages.
We chose a random $0.8-0.1-0.1$ split for train-val-test with respect to language for ($79$ languages, $9$ languages, $10$ languages), and the number of samples is $4395$, $549$, $549$ for meta-train, meta-validation, and meta-test respectively. There is no overlap between the data used for source classification, target classification, and alignment tasks.

\textbf{Hyperparameters:} We show the hyperparameters used in Table~\ref{tts_params}.

\vspace{-1mm}
\section{Additional Results}
\label{results_supp}
\vspace{-1mm}

Here we present some additional results, ablation studies, observations, and analysis on our approach.

\begin{figure*}[tbp]
\centering
\vspace{-0mm}
\includegraphics[width=\linewidth]{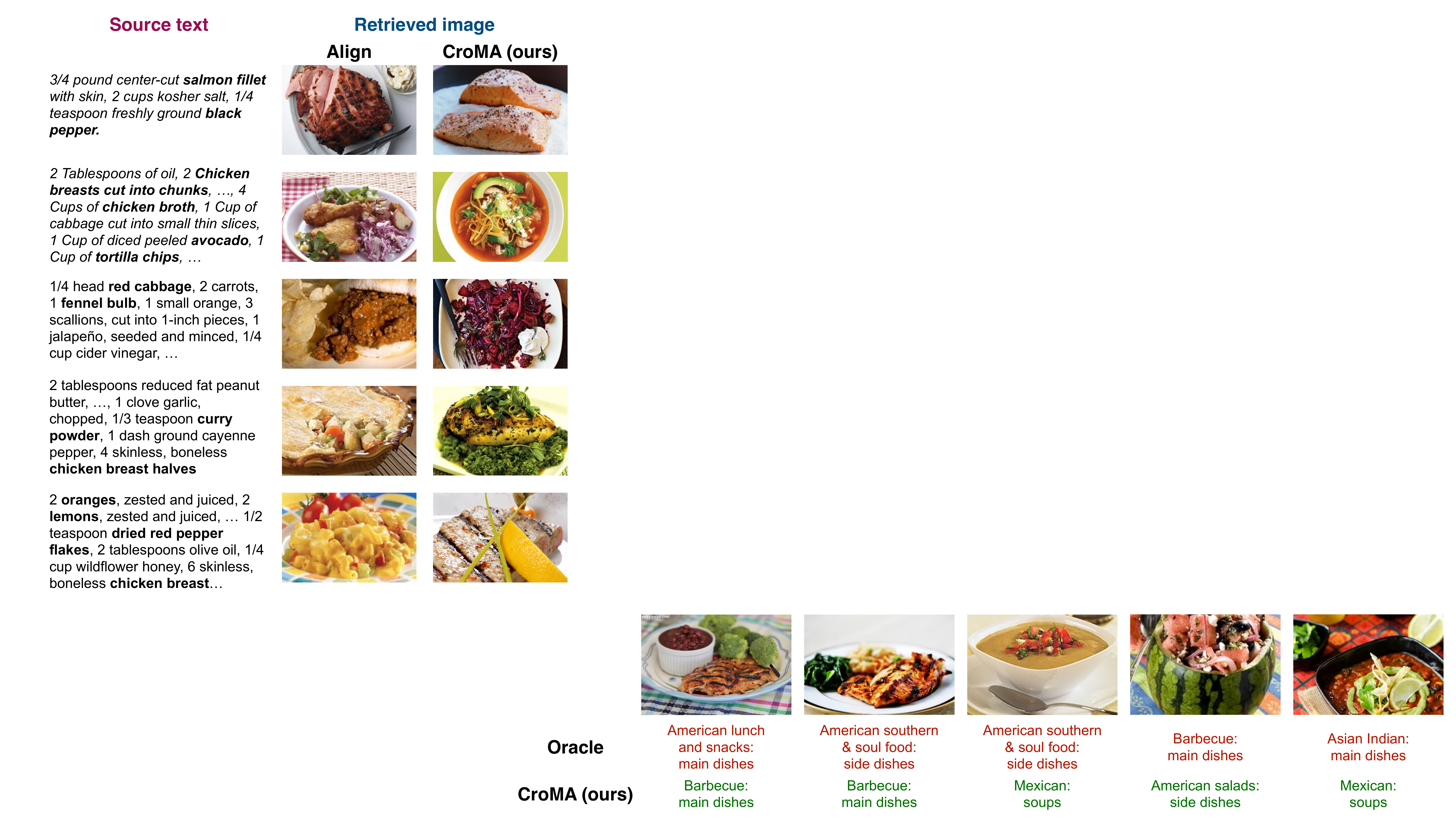}
\vspace{-6mm}
\caption{On Yummly-28K dataset, \names\ leverages source text modality to make accurate few-shot predictions on target image modality despite only seeing $1-10$ labeled image examples.\vspace{-4mm}}
\label{examples_supp}
\end{figure*}

\begin{figure}[tbp]
\centering
\includegraphics[width=\linewidth]{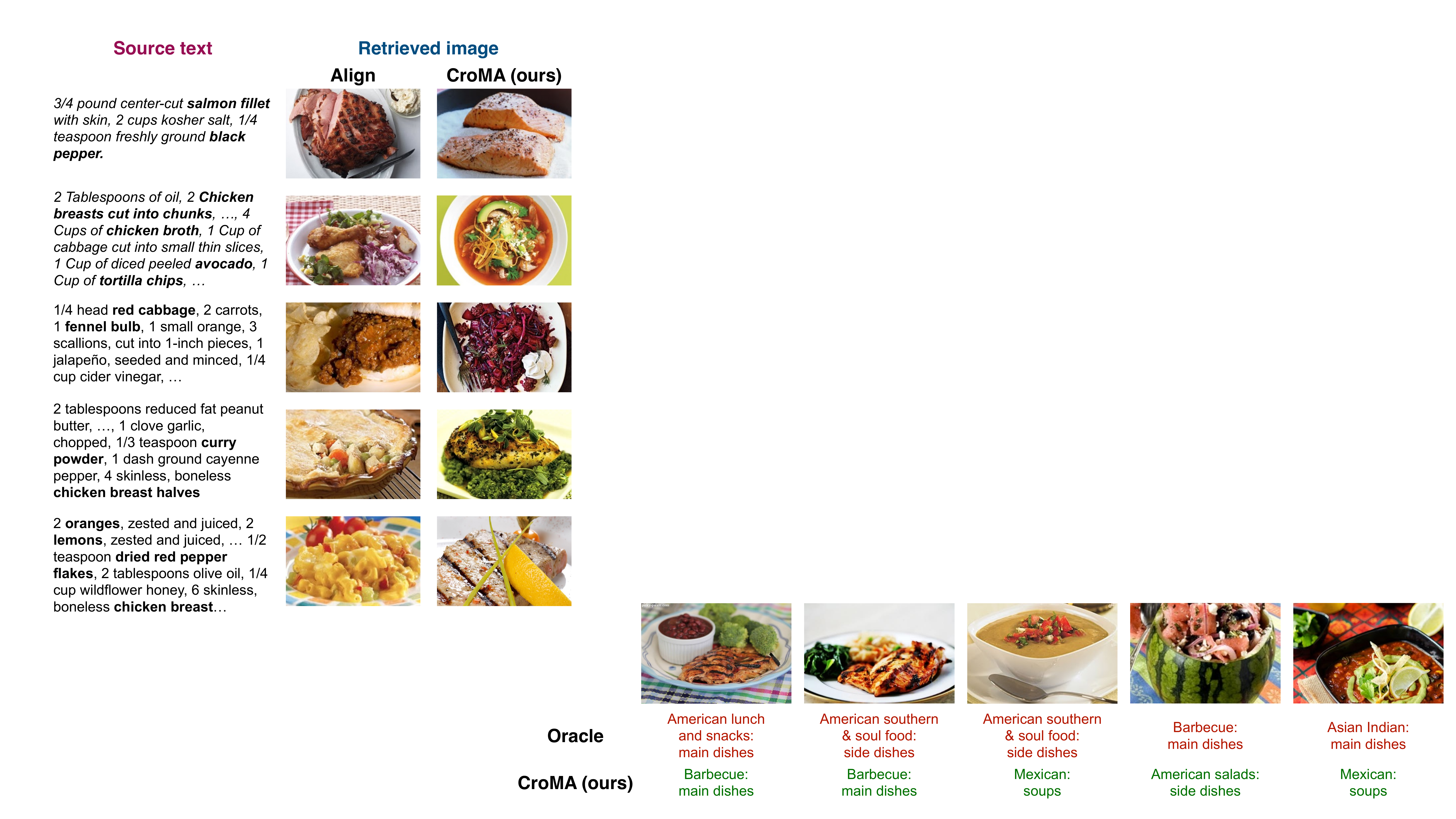}
\vspace{-2mm}
\caption{More samples of retrieved images given text recipes. \names\ performs few-shot retrieval of images more accurately than existing alignment approaches.\vspace{-4mm}}
\label{pairs_supp}
\end{figure}

\vspace{-1mm}
\subsection{Text to Image}
\vspace{-1mm}

\textbf{Extra results:} We show some samples of image to recipe label predictions on low-resource image samples in Figure~\ref{examples_supp}. Despite seeing just $5$ labeled image samples, \names\ is able to quickly generalize and recognizes images from new recipes.

In Figure~\ref{pairs_supp}, we show more samples of retrieved data in the target given input in the source modality to help us understand which source modalities the model is basing its target predictions on. Our model yields better retrieval performance than the baselines, indicating that meta-alignment successfully aligns new concepts in low-resource target modalities.

\vspace{-1mm}
\subsection{Image to Audio}
\vspace{-1mm}

\textbf{Extra results:} We implement one more baseline derived as variations from existing work and adapted to our cross-modal generalization task. We adapt unsupervised meta-learning~\cite{DBLP:journals/corr/abs-1810-02334} which uses the aforementioned $17$ weak clusters as prediction targets for the target modality during meta-training. This gives more discriminative training signal than the self-supervised reconstruction objective discussed in main text while still not explicitly using target modality labels during meta-training. We show these results in Table~\ref{image_audio_weak}. While this baseline does do better than the reconstruction version of unsupervised meta-learning, it still underperforms as compared to \name.

\vspace{-1mm}
\subsection{Text to Speech}
\vspace{-1mm}

\textbf{Extra results:} We present some extra results by comparing with existing baselines in domain adaptation and transfer learning (see Section~\ref{vs_domain}) in Table~\ref{tts_da}. We observe that none of them perform well on cross-modal generalization. Although domain confusion and alignment do improve upon standard encoder sharing, they still fall short of our approach. This serves to highlight the empirical differences between cross-modal generalization and domain adaptation. Therefore, we conclude that \textbf{1. separate encoders} and \textbf{2. explicit alignment} is important for cross-modal generalization and distinguishes it from domain adaptation.

\vspace{-1mm}
\section{Cross-modal Generalization vs Domain Adaptation}
\vspace{-1mm}

In this section we make both methodological and empirical comparisons with a related field of work, domain adaptation.

\vspace{-1mm}
\subsection{Methodological Differences}
\vspace{-1mm}

At a high-level, the core differences between cross-modal generalization and domain adaptation lies in the fact that domain adaptation assumes that both source and target data are from the same modality (e.g. image-only). As a result, these models are able to share encoders for both source and target domains~\cite{DBLP:journals/corr/TzengHZSD14}. This makes the alignment problem straightforward for this simplified version of the problem.

By sharing encoders, these domain adaptation methods do not directly model $p(x_s,x_t|m_s,m_t)$ for two different modalities, which does not provide the generalization guarantees we derived in Proposition~\ref{prop: strong align}. Without alignment, and domain adaptation is unlikely to work well since $p(x_s,x_t|m_s,m_t)$ is not modeled directly except on a few anchor points that some methods uses explicitly \cite{zhang2019category}. On the other hand, our approach explicitly models $p(x_s,x_t|m_s,m_t)$ using meta-alignment which in turn provides the guarantees in Proposition~\ref{prop: strong align}, thereby helping cross-modal generalization to low-resource modalities and tasks.

\vspace{-1mm}
\subsection{Empirical Differences}
\label{vs_domain}
\vspace{-1mm}

To further emphasize these methodological differences, we modify several classical domain adaptation methods for our task to verify that it is indeed necessary to use separate encoders and perform explicit alignment for cross-modal generalization. In particular, we implement the following baselines:

1. \textbf{Shared:} We share encoders for both modalities as much as much possible. The only non-shared parameter is a linear layer that maps data from the target modality's input dimension to the source so that all subsequent encoder layers can be shared. This reflects classical work in domain adaptation and transfer learning~\cite{DBLP:journals/corr/HuhAE16,tzeng2017adversarial}.

2. \textbf{Shared + Align:} We share encoders for both modalities and further add our alignment loss (contrastive loss) on top of the encoded representations, in a manner similar to our meta-alignment model (a similar reference in the domain adaptation literature would be~\cite{jawanpuria2020geometry}).

3. \textbf{Shared + Domain confusion:} We share encoders for both modalities and further add a domain confusion loss on top of the encoded representations~\cite{DBLP:journals/corr/TzengHZSD14}.

4. \textbf{Shared + Target labels:} Finally, we share encoders for both modalities and also use target modality labels during meta-training, in a manner similar to supervised domain adaptation~\cite{DBLP:journals/corr/abs-0907-1815}.

\textbf{Results:} We present these results in Table~\ref{tts_da} and observe that none of them perform well on cross-modal generalization. Although domain confusion and alignment do improve upon standard encoder sharing, they still fall short of our approach. Our method also outperforms the Shared + Target labels baseline which uses target modality labels to train the shared encoder during meta-training. This serves to highlight the empirical differences between cross-modal generalization and domain adaptation. Therefore, we conclude that \textbf{1. separate encoders} and \textbf{2. explicit alignment} is important for cross-modal generalization which distinguishes it from domain adaptation.

\begin{table*}[t]
\fontsize{9}{11}\selectfont
\centering
\caption{Performance on image to audio concept classification from CIFAR-10 and CIFAR-100 to ESC-50. \name\ is on par with the oracle few-shot audio baseline that has seen a thousand of labeled audio samples and outperforms existing unimodal and cross-modal baselines. \#Audio (labeled) denotes the number of audio samples and labels used during meta-training.}
\setlength\tabcolsep{2.0pt}
\begin{tabular}{l l || c c c c c c c}
\Xhline{3\arrayrulewidth}
{\sc Type} & {\sc Approach}  & {\sc 1-Shot} & {\sc 5-Shot} & {\sc 10-Shot} & {\sc \#Audio (labeled)}\\
\Xhline{0.5\arrayrulewidth}
\multirow{3}{*}{Uni} & Unsup. pre-training~\cite{baevski2019effectiveness,devlin2018bert} & $44.2 \pm 0.8$ & $72.3 \pm 0.3$ & $77.4 \pm 1.7$ & $0(0)$\\
& Unsup. meta-learning~\cite{DBLP:journals/corr/abs-1810-02334} (reconstruct) & $36.3 \pm 1.8$ & $67.3 \pm 0.9$ & $76.6 \pm 2.1$ & $920 (0)$ \\
& Unsup. meta-learning~\cite{DBLP:journals/corr/abs-1810-02334} (weak labels) & $45.6 \pm 1.3$ & $74.2 \pm 0.3$ & $83.7 \pm 0.1$ & $920 (0)$ \\
\Xhline{0.5\arrayrulewidth}
\multirow{3}{*}{Cross} & Align + Classify~\cite{cicek2019unsupervised,DBLP:journals/corr/abs-1711-03213,DBLP:journals/corr/RajNT15a,tzeng2017adversarial,10.5555/2283516.2283652} & $45.3 \pm 0.8$ & $73.9 \pm 2.1$ & $78.8 \pm 0.1$ & $920 (0)$ \\
& Align + Meta Classify~\cite{sahoo2019metalearning} & $47.2 \pm 0.3$ & $77.1 \pm 0.7$ & $80.4 \pm 0.0$ & $920 (0)$ \\
& \textbf{\names\ (ours)} & $\mathbf{47.5 \pm 0.2}$ & $\mathbf{85.9 \pm 0.7}$ & $\mathbf{92.7 \pm 0.4}$ & $920 (0)$ \\
\Xhline{0.5\arrayrulewidth}
\multirow{1}{*}{Oracle} & Within modality generalization~\cite{finn2017model,nichol2018reptile} & $45.9 \pm 0.2$ & $89.3 \pm 0.4$ & $94.5 \pm 0.3$ & $920 (920)$ \\
\Xhline{3\arrayrulewidth}
\end{tabular}
\label{image_audio_weak}
\vspace{-0mm}
\end{table*}

\definecolor{gg}{RGB}{15,125,15}
\definecolor{rr}{RGB}{190,45,45}

\begin{table*}[t]
\fontsize{9}{11}\selectfont
\centering
\caption{Performance on text to speech generalization on the Wilderness dataset. We compare \name\ with some standard domain adaptation baselines and observe that none of them perform well on cross-modal generalization. Although domain confusion and alignment do improve upon standard encoder sharing, they still fall short of our approach. This serves to highlight the empirical differences between cross-modal generalization and domain adaptation.}
\setlength\tabcolsep{2.0pt}
\begin{tabular}{l l || c c c c c c}
\Xhline{3\arrayrulewidth}
{\sc Type} & {\sc Approach}  & {\sc 1-Shot} & {\sc 5-Shot} & {\sc 10-Shot} & {\sc \#Speech (labeled)}\\
\Xhline{0.5\arrayrulewidth}
\multirow{4}{*}{Domain Adaptation} & Shared & $55.6 \pm 10.2$ & $75.2 \pm 8.4$ & $81.9 \pm 3.9$ & $4395 (0)$\\
& Shared + Align~\cite{jawanpuria2020geometry} & $59.7 \pm 7.6$ & $78.4 \pm 6.2$ & $84.3 \pm 1.5$ & $4395 (0)$\\
& Shared + Domain confusion~\cite{DBLP:journals/corr/TzengHZSD14} & $59.5 \pm 7.2$ & $76.3 \pm 9.4$ & $83.9 \pm 1.8$ & $4395 (0)$\\
& Shared + Target labels~\cite{DBLP:journals/corr/abs-0907-1815} & $57.3 \pm 9.3$ & $76.2 \pm 8.4$ & $84.0 \pm 1.9$ & $4395 (4395)$\\
\Xhline{0.5\arrayrulewidth}
\multirow{3}{*}{Cross-modal} & Align + Classify~\cite{cicek2019unsupervised,DBLP:journals/corr/abs-1711-03213,DBLP:journals/corr/RajNT15a,tzeng2017adversarial,10.5555/2283516.2283652} & $61.1 \pm 6.0$ & $74.8 \pm 2.1$ & $86.2 \pm 0.7$ & $4395 (0)$\\
& Align + Meta Classify~\cite{sahoo2019metalearning} & $65.6 \pm 6.1$ & $89.9 \pm 1.5$ & $93.0 \pm 0.5$ & $4395 (0)$\\
& \textbf{\names\ (ours)} & $\mathbf{67.9 \pm 6.6}$ & $\mathbf{90.6 \pm 1.5}$ & $\mathbf{93.2 \pm 0.2}$ & $4395 (0)$\\
\Xhline{0.5\arrayrulewidth}
\Xhline{3\arrayrulewidth}
\end{tabular}
\label{tts_da}
\vspace{-2mm}
\end{table*}

\section{On Weak Alignment}
\label{tradeoff_supp}




This section discusses some mathematical guidelines on applying our method. Future theoretical work will be directed at formalizing the discussion in this section. For example, rigorous bounds on the minimizers can be derived when the models used are Lipschitz-continuous.

We focus on providing a understanding weak alignment before extending the analysis to cover the case of strong alignment. Let $S$ denote the total number of weak-alignment sets, each with $\rho^2$ inner-set variance, and $N_t$ be the number of target data points with supervision, then, clearly, a \textit{tradeoff} in $\rho^2$ and $\frac{1}{N_t}$ exists: direct supervised learning results in a generalization error proportional to $\frac{1}{N_t}$, while weak supervision results in error proportional to $\rho^2$. Dividing $N$ data points into $S$ nearest neighbor sets, the resulting sets each have roughly $N/S$ data points. If the original data points are drawn from a uniform distribution, then, each set will have variance proportional to $\frac{1}{S}$. Then, performing weak alignment is better than doing supervised learning if
\begin{equation}
    \frac{c_s}{S} < \frac{c_t}{N_t}
\end{equation}
for some architecture and task dependent constants $c_s,\ c_t$. This means that if the number of anchored sets is large or when the number of supervised data point is very small, then one should opt for using weak-alignment.

We can also rewrite this in terms of the number of data points for each set $N_s$ we have. Since $S$ is the number of anchor points, one expects that the error in alignment decreases as $\frac{1}{S}$. Let $N=SN_s$ denote the total number of data points. 
for some architecture and task dependent constant $c_s,\ c_t$. The above inequality is equivalent to
\begin{equation}
    \frac{N_t}{S} = \frac{N_s N_t}{N } < c,
\end{equation}
for some constant $c$. If we keep both the number of datapoints in each set and the supervised datapoints constant, then the trade-off depends only on $N$. If the number of total datapoints is large, one should use weak-alignment. What is the difference between learning with strong alignment and weak alignment? Intuitively, one would expect the generalization error to vanish when $N\to \infty$ for strong alignment, since the perfect one-to-one mapping between the target and the source can be discovered in this case. For weak alignment, however, one does not achieve vanishing generalization error in principle, since a fundamental uncertainty of order $\rho^2$ exists regarding the pairing relationship between different points within a given pair of anchored sets even if $N \to\infty$.
\vspace{-1mm}
\subsection{How to Choose $S$?}
\vspace{-1mm}

In the previous section, we assumed that the center for each set is known. However, it might come as a problem in practice  if the sets are not given \textit{a priori} and if one has to resort to clustering methods such as $k-$means for finding the desired sets and estimating their centers. In this case, one has fix $N$, but variable $S$ and $N_s$. The error in alignment now depends on both $S$ and $N_s$: (1) as $S$ gets small, then the error, as discussed in the previous section, increases as $\frac{1}{S}$; (2) smaller $N_s$ makes it harder for us to estimate the center of each set, and the by the law of large numbers, we can estimate the center at error of order $\frac{1}{N_s}$. This incurs an error of order \[\frac{c_1}{S} + \frac{c_2}{N_s}= \frac{c_1}{S} +  \frac{c_2 S}{N} > 0\]
for some constants $c_1,\ c_2$. One can take derivative to find the optimal $S^*$ such that the error is minimized:
\begin{equation}
    -\frac{c_1}{S^2} + \frac{c_2}{N} = 0 \to S^* = \sqrt{\frac{c_1 N}{c_2}},
\end{equation}
i.e. $S^*$ should scale with $\sqrt{N}$.

\end{document}